\documentclass[onecolumn]{IEEEtran}
\usepackage{graphics,color}
\usepackage{subfigure}
\usepackage{epsfig}
\usepackage{times,graphicx,amssymb,amsfonts,amsmath,bm}
\usepackage{cite}
\usepackage{amsmath}
\usepackage{multirow}
\usepackage{epsfig}
\usepackage{amsmath}
\usepackage{multirow}

\newcommand{\vc}[1]{\ensuremath{\mathrm{vec}\left( #1 \right)}}

\newcommand{\alphab}{\mbox{\boldmath$\alpha$}}
\newcommand{\ab}{\ensuremath{\mathbf{a}}}

\newcommand{\Psib}{\mbox{\boldmath$\Psi$}}
\newcommand{\Gammab}{\mbox{\boldmath$\Gamma$}}

\newcommand{\E}{\mathbb{E}}
\newcommand{\Pb}{\mathbb{P}}
\newcommand{\A}{{\cal{S}}}

\newtheorem{theorem}{Theorem}

\newtheorem{lemma}{Lemma}
\newtheorem{cor}{Corollary}

\usepackage{cite}

\begin{document}

\nopagebreak
%\title{Conditions for Group Sparse Brain Network Estimation}
\title{Causal Network Inference via \\ Group Sparse Regularization}

\author{Andrew~Bolstad,~\IEEEmembership{Member,~IEEE,}
        Barry~Van~Veen,~\IEEEmembership{Fellow,~IEEE,}
        and~Robert~Nowak,~\IEEEmembership{Fellow,~IEEE}% <-this % stops a space
\thanks{This work supported in part by the NIBIB under NIH awards EB005473, EB009749 and by the AFOSR award FA9550-09-1-0140.}
\thanks{This work is sponsored by the department of the Air Force under contract FA8721-05-C-0002. Opinions, interpretations, conclusions and recommendations are those of the author and are not necessarily endorsed by the United States Government.}
}% <-this % stops a space
%\thanks{Manuscript received April 19, 2005; revised January 11, 2007.}}

%\markboth{IEEE Transactions on Signal Processing,~Vol.~XX, No.~XX, January~2010}%
%{Bolstad \MakeLowercase{\textit{et al.}}: Title}

\maketitle

\begin{abstract}
%\boldmath
  This paper addresses the problem of inferring sparse causal networks modeled by multivariate auto-regressive (MAR) processes.  Conditions are derived under which the Group Lasso (gLasso) procedure consistently estimates sparse network structure.  The key condition involves a ``false connection score'' $\psi$.  In particular, we show that consistent recovery is possible even when the number of observations of the network is far less than the number of parameters describing the network, provided that $\psi < 1$.  The false connection score is also demonstrated to be a useful metric of recovery in non-asymptotic regimes. The conditions suggest a modified gLasso procedure which tends to improve the false connection score and reduce the chances of reversing the direction of causal influence.  Computational experiments and a real network based electrocorticogram (ECoG) simulation study demonstrate the effectiveness of the approach.
\end{abstract}

%\begin{IEEEkeywords}
%causal inference, sparsity, network inference, lasso
%\end{IEEEkeywords}

% \fi
%\IEEEpeerreviewmaketitle

\section{Introduction}

The problem of inferring networks of causal relationships arises in biology, sociology, cognitive science and engineering.  Specifically, suppose that we are able to observe the dynamical behaviors of $N$ individual components of a system and that some, but not necessarily all, of the components may be causally influencing each other.  We will refer to such a system as a causal network.  To emphasize the network-centric viewpoint,
we will use the terms node and network, instead of component and system, respectively. Causal network inference is the process of identifying the significant causal influences by observing the time-series at the nodes.  %
For example, in electrocorticography (ECoG) the electrical signals in the brain are recorded directly and a goal is to identify the direction of information flow from one brain region to another.

One common tool for modeling causal influences is the multivariate autoregressive (MAR) model \cite{baccala:01,kaminski:05,winterhalder:05}.  MAR models assume that the current measurement at a given node is a linear combination of the previous $p$ measurements at all $N$ nodes, plus an innovation noise:

\begin{equation}
\label{eq:mar}
\mathbf{x}(t) = \sum_{r=1}^p \mathbf{A}_r \mathbf{x}(t-r) + \mathbf{u}(t)
\end{equation}

\noindent
where $\mathbf{x}(t) = \begin{bmatrix} x_1(t) & x_2(t) & \ldots & x_N(t) \end{bmatrix}^T$ is a vector of signal measurements across all $N$ nodes at time $t$, matrices $\mathbf{A}_r = \{ a_{i,j}(r) \}$ contain autoregressive coefficients describing the influence of node $j$ on node $i$ at a delay of $r$ time samples, and $\mathbf{u}(t) = \begin{bmatrix} u_1(t) & u_2(t) & \ldots & u_N(t) \end{bmatrix}^T \sim \mathcal{N}(\mathbf{0},\mathbf{\Sigma})$ is innovation noise.  The MAR model is especially conducive to the assessment of Granger Causality, where time series $x_j$ is said to Granger-cause $x_i$ if knowledge of the past of $x_j$ improves the prediction of $x_i$ compared to using only the past of $x_i$ \cite{lutkepohl:91}.

The MAR model in Eq.~(\ref{eq:mar}) allows for the possibility of a fully connected network in which every node causally influences every other node.  This flexibility is somewhat unrealistic and leads to practical challenges.  In many networks each node is directly influenced by only a small subset of other nodes.  The MAR model is overparameterized in such cases.  This leads to serious practical problems.   It may be impossible to reliably infer the network from noisy, finite-length time-series because of the large number of unknown coefficients in overparameterized models.  We define the Sparse MAR Time-series (SMART) model to have the same form as Eq.~(\ref{eq:mar}) but include an extra parameter ${\cal S}_{\mbox{\tiny active}}$ denoting the index pairs of non-zero causal influences to eliminate overparameterization.  For example, if node $j$ influences node $i$, then $(i,j) \in {\cal S}_{\mbox{\tiny active}}$, otherwise $(i,j) \not \in {\cal S}_{\mbox{\tiny active}}$ and $a_{i,j}(r)=0$ for all time indices $r$.  The SMART model for node $i$ is given by:

%\begin{equation}
%\label{eq:smart}
%x_i(t) = \sum_{r=1}^p \ \sum_{j:(i,j) \in \A_{active}} a_{i,j}(r) x_j(t-r) + u_i(t)
%\end{equation}

\begin{equation}
\label{eq:rearrange}
x_i(t) = u_i(t) + \sum_{j: (i,j)\in {\A}_{\mbox{\tiny active}}} \sum_{r=1}^p a_{i,j}(r) x_j(t-r)
\end{equation}

\noindent
Applying Eq.~(\ref{eq:rearrange}) to each node $i=1,2,\ldots,N$ in turn gives the SMART model for the whole network.

If the cardinality of the active set, denoted $|{\cal S}_{\mbox{\tiny active}}|$, is equal to $N^2$, then the SMART model is equivalent to the MAR model.  We are primarily interested in networks for which $|{\cal S}_{\mbox{\tiny active}}| \leq m N$, for some constant $m>1$.  In such cases, the main inference challenge is reliably identifying the set ${\cal S}_{\mbox{\tiny active}}$, since once this is done the task of estimating the SMART coefficients is a simple and classical problem.  In general, the amount of data required to reliably estimate SMART coefficients decreases as $|{\cal S}_{\mbox{\tiny active}}|$ decreases.

%Identifying ${\cal S}_{\mbox{\tiny active}}$ is a subset selection problem.  Simple subset selection problems can be solved using the well-known Lasso procedure.  The Lasso mixes an $\ell_2$ norm on the residual error with an $\ell_1$ norm penalty on the regression coefficients favoring a solution in which most coefficients are zero \cite{tibshirani:96}.  However, ordinary Lasso does not capture the group structure of sparse connections in the SMART model.  The Group Lasso (gLasso) procedure was first proposed by \cite{yuan:06} in a general setting to promote group-structured sparsity patterns.   gLasso penalties  have recently been proposed for identifying interaction patterns in the human brain \cite{haufe:08,haufe:09}, identifying gene regulatory networks (LOZANO!!!), as well as source localization in magneto-/electroencephalography (M/EEG) \cite{akb:07,akb:ssp07,ding_he:07,ou:08,haufe:08,akb_ni:09}.  We study the application of gLasso to SMART model estimation (which we term the SMART gLasso or SG) in Section~\ref{GSMAR:bckgd} by penalizing the sum of $\ell_2$ norms of the coefficients of each network link ($\ell_1$ norm of $\ell_2$ norms).

Identifying ${\cal S}_{\mbox{\tiny active}}$ is a subset selection problem.  Simple subset selection problems can be solved using the well-known Lasso procedure.  The Lasso mixes an $\ell_2$ norm on the residual error with an $\ell_1$ norm penalty on the regression coefficients favoring a solution in which most coefficients are zero \cite{tibshirani:96}.  However, ordinary Lasso does not capture the group structure of sparse connections in the SMART model.  The Group Lasso (gLasso) procedure was first proposed by \cite{yuan:06} in a general setting to promote group-structured sparsity patterns.  gLasso penalties have recently been proposed for source localization in magneto-/electroencephalography (M/EEG) \cite{akb:07,akb:ssp07,ding_he:07,ou:08,haufe:08,akb_ni:09}, as well as for identifying interaction patterns in the human brain \cite{haufe:09} and in gene regulatory networks \cite{lozano:bio09}.  In both \cite{haufe:09} and \cite{lozano:bio09} the gLasso is effectively applied to SMART model estimation by penalizing the sum of $\ell_2$ norms of the coefficients of each network link ($\ell_1$ norm of $\ell_2$ norms).  We study estimation consistency of this technique which we term the SMART gLasso or SG.

%gLasso penalties  have recently been proposed for identifying interaction patterns in the human brain \cite{haufe:08,haufe:09}, identifying gene regulatory networks (LOZANO!!!), as well as source localization in magneto-/electroencephalography (M/EEG) \cite{akb:07,akb:ssp07,ding_he:07,ou:08,haufe:08,akb_ni:09}.  We study the application of gLasso to SMART model estimation (which we term the SMART gLasso or SG) in Section~\ref{GSMAR:bckgd} by penalizing the sum of $\ell_2$ norms of the coefficients of each network link ($\ell_1$ norm of $\ell_2$ norms).

%  A number of methods exist for solving the gLasso optimization problem.

Our main contribution is a novel characterization of the special conditions needed for consistency of the SG.  These conditions are described in Section~\ref{GSMAR:sufcond}.  Existing gLasso consistency results do not apply to the temporal structure in the SMART model.  The SG consistency conditions are similar in spirit to the standard ``incoherence'' conditions encountered in the analysis of Lasso and its variants \cite{vandeGeer:09}, but are fundamentally different because of the autoregressive structure of our model.  We define the ``false connection score'' and show that it yields a condition for consistent estimation of the underlying SMART sparsity.  If this score is below one, then the network connectivity pattern can be recovered with high probability in the limit as the size of the network and the number of samples tends to infinity (although the number of samples can grow much slower than the network size).  Conversely, if this score is above one, than an estimate that identifies all the correct connections will also include at least one false positive with high probability.

We also propose a variant of the SG in Section~\ref{GSMAR:bckgd} which does not penalize self-connections (i.e., each node is free to influence itself).  We call this variant Self-Connected SMART gLasso (SCSG) and show that it typically results in a lower false connection score for SMART models.  We provide some example networks as well as their false connection scores for the SMART gLasso and SCSG approaches in Sec.~\ref{GSMAR:samplenets}.  We demonstrate the effectiveness of our results by simulating a variety of networks in Sec.~\ref{GSMAR:sims}.  We also apply our results to a realistic brain network in Sec.~\ref{macaque} by simulating the sparse connectivity pattern observed in the macaque brain.

\section{Graph Inference with Lasso-Type Procedures}
\label{GSMAR:bckgd}

In this section we introduce the Lasso, gLasso, SG, and SCSG, and discuss previous consistency results.

\subsection{Lasso and gLasso}

Tibshirani first proposed the Least Absolute Shrinkage and Selection Operator (Lasso) in 1996 to ``retain the good features of both subset selection and ridge regression'' \cite{tibshirani:96}.  Although originally stated as an $\ell_1$ norm constrained least squares optimization, the Lasso can also be stated as an unconstrained mixed-norm minimization.  We consider the unconstrained problem throughout:

\begin{equation}
\label{eq:lasso_basic}
\hat{\mathbf{a}}^{Lasso} = \arg\min_{\alphab} \frac{1}{n} \| \mathbf{y} - \mathbf{X} \alphab \|_2^2 + \lambda \| \alphab \|_1
\end{equation}

\noindent
Here it is assumed that measured length $n$ vector $\mathbf{y}$ is the result of a sparse linear combination of columns of $\mathbf{X}$; i.e. $\mathbf{y} = \mathbf{X} \mathbf{a}$ for sparse vector $\mathbf{a}$.  The first term of (\ref{eq:lasso_basic}) penalizes solutions which do not fit the measured data well, while the second term favors solution which are sparse.  Yuan and Lin \cite{yuan:06} introduced the Group Lasso (gLasso) extension to Tibshirani's Lasso in 2006.  While the Lasso penalizes the $\ell_1$ norm of the coefficient vector, the gLasso divides the coefficient vector into predetermined sub-vectors and penalizes the sum of the $\ell_2$ norms of the sub-vectors; i.e., the $\ell_1$ norm of $\ell_2$ norms:

\begin{equation}
\label{eq:glasso_basic}
\hat{\mathbf{a}}^{gLasso} = \arg\min_{\alphab} \frac{1}{n} \left\| \mathbf{y} - \mathbf{X} \begin{bmatrix} \alphab_1 \\ \vdots \\ \alphab_N \end{bmatrix} \right\|_2^2 + \lambda \sum_{i=1}^N \| \alphab_i \|_2
\end{equation}

\noindent
Such a penalty is beneficial when each group of coefficients is believed to be either all zero or all non-zero, and the solution contains only a small number of nonzero coefficient groups, e.g.,  \cite{akb:07,akb:ssp07,ding_he:07,ou:08,haufe:08,akb_ni:09}.

% insert new

Solving the SMART model subset selection problem  with the gLasso leads to the SG estimate:

\begin{equation}
\label{eq:glasso_ar}
\widehat{\mathbf{a}}_i^{SG} = \arg\min_{\mathbf{a}_i} \frac{1}{n} \left\| \mathbf{y}_i - \mathbf{X} \ab_i \right\|_2^2 + \lambda \sum_{j=1}^N \| \mathbf{a}_{i,j} \|_2
\end{equation}

\noindent
where we define:

\begin{eqnarray*}
\mathbf{y}_i & = & \begin{bmatrix} x_i(t) & x_i(t-1) & \ldots & x_i(t-n+1) \end{bmatrix}^T \\
\mathbf{X}_i & = & \begin{bmatrix} x_i(t-1) & \ldots & x_i(t-p) \\ x_i(t-2) & \ldots & x_i(t-p-1) \\ \vdots & \ddots & \vdots \\ x_i(t-n) & \ldots & x_i(t-p-n+1) \end{bmatrix}  \\
\mathbf{X} & = & \begin{bmatrix} \mathbf{X}_1 & \mathbf{X}_2 & \ldots & \mathbf{X}_N \end{bmatrix} \\
\mathbf{a}_{i,j} & = & \begin{bmatrix} a_{i,j}(1) & a_{i,j}(2) & \ldots & a_{i,j}(p) \end{bmatrix}^T \\
\mathbf{a}_i & = & \begin{bmatrix} \mathbf{a}_{i,1} & \mathbf{a}_{i,2} & \ldots &\mathbf{a}_{i,N} \end{bmatrix}^T
\end{eqnarray*}

%\mathbf{X}_i & = & \begin{bmatrix} x_i(t-1) & x_i(t-2) & \ldots & x_i(t-p) \\ x_i(t-2) & x_i(t-3) & \ldots & x_i(t-p-1) \\ \vdots & \vdots & \ddots & \vdots \\ x_i(t-n-1) & x_i(t-n-1) & \ldots & x_i(t-n-p) \end{bmatrix}  \\

%\mathbf{u}_i & = & \begin{bmatrix} u_i(t_1) & u_i(t_2) & \ldots & u_i(t_n) \end{bmatrix}^T \\

%For $i=1,\dots,N$ let

%\noindent
%so that $\mathbf{y}_i$ represents the current sample (actually current samples from $n$ realizations of the process) at node~$i$, $\mathbf{X}_i$ represents past samples at node~$i$, $\mathbf{u}_i$ is driving noise, and $\mathbf{a}_{i,j}$ contains the coefficients describing the connection from node~$j$ to node~$i$.

\noindent
The SCSG removes the penalty for  self-connections, that is, each node's own past values are allowed to predict its current value without a penalty:
\begin{equation}
\label{eq:glvar}
\hat{\mathbf{a}}_i^{SCSG} = \arg\min_{\mathbf{a}_i} \frac{1}{n} \| \mathbf{y}_i - \mathbf{X} \mathbf{a}_i \|_2^2 + \lambda \sum_{j\neq i} \| \mathbf{a}_{i,j} \|_2
\end{equation}
This represents the expectation of sparse connectivity between nodes.

The gLasso optimization falls into a class of well-studied convex optimization problems.  Many algorithms have been proposed for solving this sort of problem (see \cite{wright:08} for a description and comparison of several approaches).  Greedy procedures, such as group orthogonal matching pursuit, have been proposed as well \cite{lozano:nips09}.  The choice of optimization algorithm is not an important concern in this paper; rather the main contribution of this paper is to characterize the behavior and consistency of the solution of Eqs.~(\ref{eq:glasso_ar}) and (\ref{eq:glvar}).

%Equations~\ref{eq:glasso_ar} and \ref{eq:glvar} are convex optimization problems which can be solved with a variety of gLasso solvers (see \cite{wright:08} for a partial review).  In the sequel we use an iterative shrinkage/thresholding (IST) type procedure based on our expectation-maximization (EM) algorithm \cite{akb_ni:09}.  IST approaches are competitive solvers which can efficiently calculate the regularization path for a range of $\lambda$, and we have significant experience with our implementation.  Greedy procedures, such as group orthogonal matching pursuit, have been proposed as well \cite{lozano:nips09}; however, our results do not apply directly to non-convex approaches.  The contribution of this paper is not to propose or promote a particular solver, but rather to examine the quality and properties of the solutions of Eqs.~\ref{eq:glasso_ar} and \ref{eq:glvar}.

%Many solvers have been developed to solve the gLasso convex optimization problem (see \cite{wright:08} for a partial review).  We advocate iterative shrinkage/thresholding (IST) type procedures such as our expectation-maximization (EM) algorithm \cite{akb_ni:09} and Sparse Reconstruction by Separable Approximation (SpaRSA) \cite{wright:08}.  In the sequel we use the EM approach.

%The gLasso optimization problem can be set up and solved as a second order cone program (SOCP), though much more efficient solvers have been developed, e.g. \cite{wright:08}.

\subsection{Graphical Model Identification}

Lasso-like algorithms have found application in high dimensional graphical model identification.  The seminal work in this area was done by Meinshausen and B\"{u}hlmann \cite{meins:06} who consider estimating the structure of sparse Gaussian graphical models by identifying the nonzero entries of the inverse covariance matrix.  They consider an undirected graph where each vertex represents a variable and edges represent conditional dependence between two variables given all other variables.  Conditionally independent variables do not share an edge and correspond to a zero entry in the inverse covariance matrix.  Identifying the edge set, or nonzero entries in the inverse covariance matrix, is achieved by writing independent samples of one variable as a sparse, but unknown linear combination of the corresponding samples of the other variables, then using the Lasso.  Meinshausen and B\"{u}hlmann \cite{meins:06} show that this procedure consistently identifies the edge set even when the number of variables (vertices) grows faster than the number of samples.  Ravikumar, et al., \cite{ravikumar:08} propose an alternative Lasso like approach to the same problem by maximizing the $\ell_1$ norm penalized log-likelihood function.  In this case the first term of Eq.~(\ref{eq:lasso_basic}) is replaced with an inner product and log-determinant of the covariance matrix.  The graphical lasso technique solves this type of problem efficiently for very large problems \cite{friedman:08}.

\begin{figure}[htbp]
%\setcaptionwidth{3in}
\centering
\subfigure[ SMART Model Temporal Depiction ] {\label{fig:smartnetsa}
\includegraphics[height=2in]{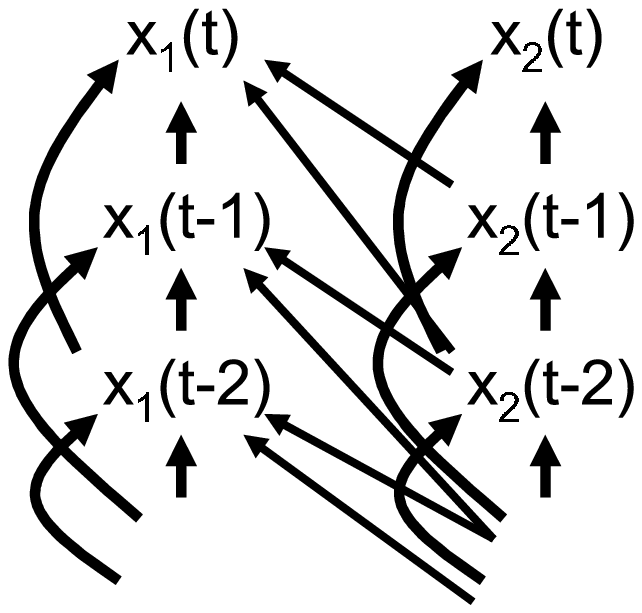}
} \qquad
\subfigure[ SMART Model Network Depiction ] {\label{fig:smartnetsb}
\includegraphics[height=0.4in]{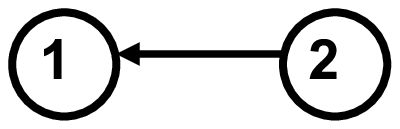}
}
\caption{Two graphical depictions of a two node, second order SMART model. (a) Explicit time dependence structure. (b)  Shorthand depiction of (a) suppressing time and self-connections.}
\label{fig:smartnets}
\end{figure}

The SMART model is a graphical model involving causal relationships and consequently, an element of time.  The resulting model is a directed graph, and each node can be represented by multiple vertices: one for the current value, and potentially infinitely many for past values at that node as shown in Fig.~\ref{fig:smartnetsa}.  To ease visualization, we suppress time dependence and illustrate causal influence with a single arrow linking one vertex per node as shown in Fig.~\ref{fig:smartnetsb}.  Here we have not shown self-connections.  Nodes which have a causal influence are termed  ``parent nodes'' (node~2 in Fig.~\ref{fig:smartnets}) and the nodes they influence  ``child nodes'' (node~1 in Fig.~\ref{fig:smartnets}).  Given that graphs representing MAR models are directed, the existing analyses by Ravikumar, et al., \cite{ravikumar:08} and Meinshausen and B\"{u}hlmann \cite{meins:06} are insufficient.  The additional notions of causality and a temporal element place the SMART model in the realm of graphical Granger models \cite{lozano:bio09,lozano:kdd09}.

% brain part
%Recently, Lasso-type procedures have been applied to MAR model estimation of brain activity.  Valdes-Sosa et al. \cite{valdes:05} apply the Lasso, as well as other sparsity inducing nonconvex techniques, to simulated and fMRI data.  The Elastic Net, essentially the Lasso plus an additional $\ell_2$ penalty, is applied to fMRI data in \cite{carroll:09}.  In 2008, Haufe, et al. \cite{haufe:09} proposed a gLasso variant for MAR network estimation and compared it to other techniques via simulation of relatively small networks, some of which we reproduce below.  They employ a gLasso variant as a regularization term in a more complicated cost function in \cite{haufe:09b}.  None of these works explore theoretical behavior of sparsity inducing techniques.

\subsection{Existing Lasso and gLasso Consistency Results}

There are many existing results on consistency of the Lasso (e.g., \cite{meins:06,zou:06}) and extensions of these to the gLasso or closely related problems (e.g., \cite{tropp_ssac:06,meier:08,cotter:05,chen_huo:04,tropp_ssag:06,obozinski:08,wang_leng:08,liu_zhang:09,lozano:nips09}).  An important concept in all these results is mutual incoherence, the maximum absolute inner product between two columns of $\mathbf{X}$.  Mutual incoherence is extended to grouped variables by using the maximum singular value of $\mathbf{X}_i^T\mathbf{X}_j$ in place of the vector inner product.  Analyzing mutual coherence in the SMART model setting is challenging due to the strong statistical dependence between columns of $\mathbf{X}$.  Both Lasso and gLasso have recently been successfully applied to SMART networks (e.g. \cite{valdes:05,haufe:09,lozano:bio09,songsiri:10}), but consistency was not considered.  In independent work, the consistency of first-order AR models (a special case of the general problem considered here) is investigated in \cite{bento:10}.  We identify novel incoherence conditions tailored specifically to the SMART model, and show how the network structure of the model affects these conditions.  Thus these incoherence conditions provide unique insight into the capabilities and limitations of SG model identification.

\section{Asymptotic Consistency of SMART gLasso}
\label{GSMAR:sufcond}

%The solution to (\ref{eq:glasso_ar}) can be sparse, due to the regularization term.  The solution $\widehat{\mathbf{a}}_i^{SG}$ is composed of subvectors or groups $\widehat{\mathbf{a}}_{i,j}^{SG}$, $j=1,\dots,N$. Some of the $\widehat{\mathbf{a}}_{i,j}^{SG}$ may be identically zero. The natural interpretation of the result $\widehat{\mathbf{a}}_{i,j}^{SG} = 0$ is that node $j$ does not have a significant causal influence on node $i$.

In this section we provide sufficient conditions for the asymptotic consistency of the SG estimate assuming the data are generated by a SMART model.  Our general approach is similar to the style of argument used in the analysis of gLasso consistency \cite{liu_zhang:09} and other graph inference methods based on sparse regression \cite{meins:06}. An important distinction in SG is the MAR structure of the design matrix $\mathbf{X}$.

%That is, $\A_i = \{j \in \{1,\dots,N\} \, : \, (i,j)\in \A_{\mbox{\tiny active}} \}$.

Let $\A_i= \{j \in \{1,\dots,N\} \, : \, (i,j)\in \A_{\mbox{\tiny active}} \}, i=1,\dots,N$ indicate the subset of nodes that causally influence node $i$.  Define $\mathbf{X}_{\A_i}$ and $\mathbf{X}_{\A_i^C}$ to be submatrices of $\mathbf{X}$ composed of the matrices $\mathbf{X}_j$, $j\in \A_i$ and $\mathbf{X}_j$, $j \not \in \A_i$, respectively.  An oracle that knows $\A_i$ does not need to solve the subset selection problem but only a regression problem with design matrix $\mathbf{X}_{\A_i}$ and parameters $\mathbf{a}_{i,j}$, $j\in \A_i$.

%An oracle that knows $\A_i$ would use only $\mathbf{X}_{\A_i}$ to estimate the coefficients of the causal influences $\mathbf{a}_{i,j}$, $j\in \A_i$.

Our main result makes use of a regression problem with the same design matrix.  Consider a node $j$ with $j \not \in \A_i$.  The optimal linear predictor of $\mathbf{X}_j$ given $\mathbf{X}_{\A_i}$ is $\sum_{k \in \A_i} \mathbf{X}_k \Psib_{j,k}$ where the $\Psib_{j,k}$ minimize $\E[ \| \mathbf{X}_j - \sum_{k \in \A_i} \mathbf{X}_k \Psib_{j,k} \|_F^2 ]$.  If we stack $\{\Psib_{j,k}\}_{k \in \A_i}$ to form a matrix $\Psib_{j,\A_i}$, then we can write $\sum_{k \in \A_i} \mathbf{X}_k \Psib_{j,k}  = \mathbf{X}_{\A_i}^T \Psib_{j,\A_i}$.  Using standard matrix calculus it is not difficult to verify that $$\Psib_{j,\A_i} =  \mathbf{R}_{\A_i,\A_i}^{-1} \E[\mathbf{X}_{\A_i}^T \mathbf{X}_j] \ $$
where the covariance matrix
$$
\mathbf{R}_{\A_i,\A_i} = \E[\mathbf{X}_{\A_i}^T \mathbf{X}_{\A_i}].
$$

%\begin{eqnarray*}
%\mathbf{R}_{\A_i,\A_i} & = & \E[\mathbf{X}_{\A_i}^T \mathbf{X}_{\A_i}] \\
%\mathbf{R}_{\A_i,\A_i^C} & = & \E[\mathbf{X}_{\A_i}^T \mathbf{X}_{\A_i^C}] \\
%\end{eqnarray*}

%\textcolor{magenta}{I think this eliminates the need for the lengthy derivation in the appendix (Appendix D).}

%We will need to assume that entries in $\mathbf{y}_i$ and the corresponding row of the $\mathbf{X}_i$ matrices come from independent realizations of the SMART network.  In practice consecutive entries in $\mathbf{y}_i$ are highly correlated by definition, but taking every

%In other words, $\mathbf{y}_i(t_k) = x_i(t_k)$ and $\mathbf{X}_i(k,\cdot) = [ \begin{smallmatrix} x_i(t_k-1) & \ldots & x_i(t_k-p) \end{smallmatrix} ]$ for independent

% $c_1$, $c_2$, $c_3$, $c_4$, 
%\item \label{as:scaling} \textbf{Scaling:} number of nodes: $N = \mathcal{O}(n^{c_1})$, maximum number of parent nodes: $m = \mathcal{O}(n^{c_2})$, model order: $p = \mathcal{O}(n^{c_3})$, and regularization parameter: $\lambda \asymp n^{-c_4/2}$ with $c_2 < c_4$ and $c_3 + c_4 < 1$.

Recall the following variables: $N$, the number of nodes in the network; $m$, the maximum number of parent nodes; $p$, the SMART model order; and $n$, the number of observations.  The main result concerning the consistency of SMART gLasso is
\begin{theorem}
\label{thm:ascon}
Let $C_{power}$, $C_{con}$, $C_{min}$, $C_{max}$, and $C_{fcs}$ be non-negative constants.  Assume entries in $\mathbf{y}_i$ and the corresponding row of each $\mathbf{X}_j$ matrix come from independent realizations of the SMART model.  Assume the following conditions hold:
\begin{enumerate}
\item \label{as:scaling} \textbf{Scaling:} $N$, $m$, and $p$ are $\mathcal{O}(n^c)$, while $\lambda$ is $\Theta(n^{-c})$ for different $c>0$ with $m \lambda^2 = o(1)$ and $\frac{p}{n \lambda^2} = o(1)$.
\item \label{as:dnoise} \textbf{Signal Power:} $$\max_{i\in\{1,\dots,N\}} \sigma_i^2 = \E[x_i^2(t)] \leq C_{power} < \infty$$
\item \label{as:cstrength} \textbf{Connection Strength:} $\min_{(i,j) \in \A_{active}} \| \mathbf{a}_{i,j} \|_2 \geq C_{con} > 0$
\item \label{as:minsv} \textbf{Minimum Power:} $\max_i \| \mathbf{R}_{\A_i,\A_i}^{-1} \|_2 \leq C_{min}^{-1} < \infty$
\item \label{as:maxsv} \textbf{Maximum Cross Correlation:} $$\max_i \| \mathbf{R}_{\A_i,\A_i^C} \|_2 \leq C_{max} < \infty$$
where
$$ \mathbf{R}_{\A_i,\A_i^C} = \E[\mathbf{X}_{\A_i}^T \mathbf{X}_{\A_i^C}] $$
\item \label{as:represent} \textbf{False Connection Score:} For all $(i,j) \in \A_{\mbox{\tiny active}}^C$

\begin{equation}
\label{eq:fcs}
\psi_{j \rightarrow i}^{FC} := \left\| \sum_{k \in \A_i} \Psib_{j,k}^T  \frac{\mathbf{a}_{i,k}}{\| \mathbf{a}_{i,k} \|_2} \right\|_2 \leq C_{fcs} < 1
\end{equation}

\end{enumerate}

\noindent
Then for all $n$ sufficiently large, the set of links identified by SG satisfies $\hat{{\cal S}}={\cal S}_{\mbox{\tiny active}}$  with probability greater than $1-\exp(-\Theta(n))$; i.e., zero and nonzero links identified by SG agree with those of the underlying true model.
\end{theorem}

\begin{IEEEproof}
The proof is presented in Appendix~\ref{GSMAR:asymptcons}.
\end{IEEEproof}

Note we have used the following notation: $f(n) = \mathcal{O}(g(n))$ implies $|f(n)| \leq k |g(n)|$ for some $k>0$ and large $n$, $f(n) = \Theta(g(n))$ implies $k_1 |g(n)| \leq |f(n)| \leq k_2 |g(n)|$ for some positive constants $k_1$ and $k_2$ and large $n$, and  $f(n) = o(g(n))$ implies $|f(n)| \leq k |g(n)|$ for all $k>0$ and large $n$.

%Furthermore, let $ \Theta(n)$ satisfy $0 \leq k_1 n \leq \Theta(n) \leq k_2 n$ for all $n \geq n_0$ where $k_1$, $k_2$ and $n_0$ are positive constants. 

Assumption~\ref{as:scaling} specifies how network parameters grow as a function of the number of observations $n$.  It may be possible to allow some or all of the constants $C_{power}$, $C_{con}$, $C_{min}$, $C_{max}$, and $C_{fcs}$ to depend on $n$, but for the purposes of this paper we will take these to be constants.  The number of nodes in the network $N$ can grow at any polynomial rate, including both faster or slower than the number of observations $n$, or remain fixed. Assumptions~\ref{as:dnoise}--\ref{as:maxsv}  are rather mild.  They are used to show that there will be no false negatives for sufficiently small $\lambda$. In practice, signals are often normalized to have equal power across nodes, which automatically achieves \ref{as:dnoise}, though only this weaker assumption is necessary here.  The effect of normalization on the other assumptions, particularly \ref{as:represent}, is an interesting open question.  Assumption~\ref{as:minsv} essentially says that each time sample in the active set contains some independent information.  Assumption~\ref{as:maxsv} ensures that any influence due to the nodes in $\A_i$ cannot be easily generated using nodes in $\A_i^c$ instead.

%We mention that it may be possible to allow $C_{min}$ and $C_{max}$ to depend on $n$, but for the purposes of this paper we will take these to be constants.

Assumption~\ref{as:represent} is the most restrictive and most informative. In the proof of the theorem, Assumption~\ref{as:represent} is used to show that the probability of declaring a nonzero connection when none exists (i.e. a false connection or false alarm) goes to zero for large $n$. In order to understand the implications of the assumption, we point out a more restrictive, but less complicated alternative: $\sum_{k \in \A_i} \| \Psib_{j,k} \|_2 \leq C_{fcs} < 1$.  If this inequality holds, Assumption~\ref{as:represent} follows from simple norm bounds.  The inequality also suggests the following interpretation of Assumption~\ref{as:represent}.  Nodes that do not directly drive the node of interest (i.e., nodes in $\A_i^C$) cannot be easily predicted from nodes that are directly driving the node of interest.  In Section~\ref{GSMAR:samplenets} we provide example networks that do and do not satisfy Assumption~\ref{as:represent} to gain insight into the nature of which networks can be recovered.  We show next that Assumption~\ref{as:represent} is necessary for a large class of networks, including those of fixed size.

\begin{theorem}
\label{thm:nec}
Suppose Assumptions~\ref{as:dnoise}--\ref{as:maxsv} of Theorem~\ref{thm:ascon} hold, but $\psi_{j \rightarrow i}^{FC} \geq 1 + c$ for some pair $(i,j)$ and constant $c>0$.  Suppose also that $m^2 p < n$ for large $n$.  Then with probability exceeding $1-\exp{(-\Theta(n))}$, the connections recovered by SG will not be the true connections.
\end{theorem}

%$\left\| \sum_{k \in \mathcal{A}} \Psib_{j,k}^T  \frac{\mathbf{a}_k}{\| \mathbf{a}_k \|_2} \right\|_2 \geq 1 + c$

\begin{IEEEproof}
A proof is given in Appendix~\ref{GSMAR:neccond}.
\end{IEEEproof}

Theorem~\ref{thm:nec} suggests that the false connection score is extremely important in sparse network recovery, especially in finite parameter networks, which are discussed below in Sec.~\ref{GSMAR:recfinite}.

%The slightly different scaling law $n>m^2 p$ for large $n$ (equivalently $2c_2 + c_3 <1$) may be an artifact of the proof technique.  Furthermore, for most practical cases, the network itself is fixed (i.e. $m$, $p$, and $N$ are constant) and only the number of samples $n$ increases.

The SCSG (\ref{eq:glvar}) assumes that each node is driven by its own past.  The conditions of Theorem~\ref{thm:ascon}, with minor modification, still govern the ability to recover the correct connectivity pattern using SCSG:

\begin{cor}
\label{cor:freeparam}
Suppose Assumptions~\ref{as:scaling}--\ref{as:maxsv} of Theorem~\ref{thm:ascon} hold for all $l$.  In place of Assumption~\ref{as:represent}, assume:

\begin{equation}
\label{eq:modrep}
\widetilde \psi_{j \rightarrow i}^{FC} = \left\| \sum_{k \in \A_i,k \neq i} \Psib_{j,k}^T  \frac{\mathbf{a}_{i,k}}{\| \mathbf{a}_{i,k} \|_2} \right\|_2 \leq C_{fcs} < 1.
\end{equation}

\noindent
Then with probability exceeding $1-\exp{(-\Theta(n))}$, the connections recovered by SCSG (\ref{eq:glvar}) will be the true connections.
\end{cor}

\begin{IEEEproof}
See Appendix~\ref{app:corollaryproof}.
\end{IEEEproof}

As we will show in the next section, $\widetilde \psi_{j \rightarrow i}^{FC}$ is typically lower than $\psi_{j \rightarrow i}^{FC}$, though cancellation between the self-connection term and other terms in the sum of (\ref{eq:fcs}) is possible.

\section{Network Recovery}

In Section~\ref{GSMAR:sufcond} we established conditions which guarantee high probability recovery of SMART networks asymptotically, allowing the network size to grow faster than the number of samples.   Next we explore the differences between the asymptotic setting and finite sample regimes.

\subsection{Recovery of Finite Parameter Networks}
\label{GSMAR:recfinite}

%\section{Necessary Condition for Recovery of Fixed Networks???}

In practice, the network parameters are typically fixed, and we are interested in performance as the number of measurements $n$ grows.  The results of Theorems~\ref{thm:ascon} and \ref{thm:nec} still apply.  In the finite network case, $m$, $p$, and $N$ are fixed, so $(m^2 p)/n$ tends to zero and Assumption~\ref{as:scaling} is satisfied as long as $\lambda^2 = \mathcal{O}(n^{-c})$ with $0 < c < 1$.  Also, Assumptions~\ref{as:dnoise}--\ref{as:maxsv} are automatically satisfied as long as there is driving noise in each node. Assumption~\ref{as:represent}  is the only one that does not necessarily hold. This implies the following corollary, which follows immediately from the proof of Theorem~\ref{thm:ascon}.

\begin{cor}
\label{cor:fixednet}
For a SMART model with fixed parameters, (\ref{eq:glasso_ar}) will recover the correct network structure with probability greater than $1-\exp{(-\Theta(n))}$ if $\psi_{j \rightarrow i}^{FC} < 1$ for all pairs $(i,j) \in \A_{\mbox{\tiny active}}^C$.  If $\psi_{j \rightarrow i}^{FC} > 1$ for some $(i,j) \in \A_{\mbox{\tiny active}}^C$, then (\ref{eq:glasso_ar}) will fail to recover the correct structure with probability exceeding $1-\exp{(-\Theta(n))}$.  The same result holds for (\ref{eq:glvar}) using $\tilde{\psi}_{j \rightarrow i}^{FC}$.
\end{cor}

\subsection{Recovery of Known Networks}

Given Corollary~\ref{cor:fixednet} it is easy to check whether a  given SMART model structure can be recovered via (\ref{eq:glasso_ar}) or (\ref{eq:glvar}).  Define $\Gammab(\tau) = \E[ \mathbf{x}(t) \mathbf{x}^T(t-\tau) ]$, and recall $\mathbf{\Sigma}$ is the driving noise $\mathbf{u}(t)$ covariance matrix.  If we define the collection of MAR coefficients $\mathbf{A}$ and $\tilde{\mathbf{\Sigma}}$ as:
\begin{eqnarray*}
\mathbf{A} & = & \begin{bmatrix} \begin{matrix} \mathbf{A}_1 & \mathbf{A}_2 & \ldots \end{matrix} & \mathbf{A}_p \\ \mathbf{I}_{N(p-1)} & \mathbf{0}_{N(p-1),N} \end{bmatrix}, \\
\tilde{\mathbf{\Sigma}} & = & \begin{bmatrix} \mathbf{\Sigma} & \mathbf{0}_{(p-1)N} \\ \mathbf{0}_{(p-1)N} & \mathbf{0}_{(p-1)N} \end{bmatrix},
\end{eqnarray*}

\noindent
then $\Gammab(\tau)$ can be calculated via (see e.g. \cite{lutkepohl:91})

\begin{equation}
\label{eq:cov}
\Gammab = \mathbf{A} \Gammab \mathbf{A}^{T} + \tilde{\mathbf{\Sigma}}
\end{equation}

\noindent
where

$$
\Gammab = \begin{bmatrix} \Gammab(0) & \Gammab(1) & \ldots & \Gammab(p-1) \\ \Gammab(-1) & \Gammab(0) & \ldots & \Gammab(p-2) \\ \vdots & \vdots & \ddots & \vdots \\ \Gammab(1-p) & \Gammab(2-p) & \ldots & \Gammab(0) \end{bmatrix}.
$$

\noindent
Using properties of Kronecker products, (\ref{eq:cov}) can be solved in closed form:

\begin{equation}
\label{eq:calccov}
\vc{\Gammab} = (\mathbf{I} - \mathbf{A} \otimes \mathbf{A})^{-1} \vc{\tilde{\mathbf{\Sigma}}}.
\end{equation}

Given this closed form expression for $\Gammab$, matrices $\mathbf{R}_{\mathcal{S}_i,\mathcal{S}_i}$ and $\mathbf{R}_{\mathcal{S}_i,\mathcal{S}_i^C}$ are formed for each node $i$ by selecting the appropriate entries from covariance matrix $\Gammab$ and subsequently used to calculate $\Psib_{j,\A_i}$.  Given $\Psib_{j,\A_i}$ and $\mathbf{a}_{i,k}$ for all $k \in \A_i$, $\psi_{j \rightarrow i}^{FC}$ or $\tilde{\psi}_{j \rightarrow i}^{FC}$ can be calculated and compared to one via Eq.~(\ref{eq:fcs}) or (\ref{eq:modrep}), respectively.

\subsection{Challenges in Realistic Networks}

The theoretical basis for SMART model recovery relies on independent data samples and asymptotic probability concentration arguments.  We now consider consequences of more realistic data sets.

Our analysis focuses on the dependence accross columns of $\mathbf{X}$ and the corresponding entry of $\mathbf{y}_i$ induced by the SMART model.  To prove Theorems~\ref{thm:ascon} and \ref{thm:nec}, we assumed each row of $\mathbf{X}$ and the corresponding entry of $\mathbf{y}_i$ to be independent from other rows.  This is not true in realistic networks where each $\mathbf{X}_i$ is actually Toeplitz; however, rows of $\mathbf{X}$ and $\mathbf{y}_i$ decorrelate as the time lag between them grows ($\E [ \mathbf{x}(t) \mathbf{x}^T(t-\tau) ] \approx \mathbf{0}$).  The simulations in Secs.~\ref{GSMAR:sims} and \ref{macaque} use correlated rows and reveal that the false alarm score has a more significant impact on performance than the row dependence.  The effect of row dependence has been consider in the special case of first order (p = 1) AR models in [33], which yields a lower bound on the required number of observations.

%Incorporating analysis of row dependence into our SMART model consistency results is an important and challenging problem.  Bento, et al., \cite{bento:10} analyze a first order ($p=1$) SMART model (and corresponding continuous time model) and provide a lower bound on the observation time necessary for network recovery in a recent preprint.  Merging this work with our false connection score (for the $p>1$ case) is interesting future work which could provide a comprehensive picture of SMART network recovery.

%To prove Theorems~\ref{thm:ascon} and \ref{thm:nec}, we assumed independent samples comprised each row of $\mathbf{X}$ and each entry of $\mathbf{y}_i$. Since signals generated by MAR models are, by their very nature, dependent on past values, each new data vector $\mathbf{x}(t)$ will have some dependence on past data.  On the other hand, the correlation between $\mathbf{x}(t)$ and $\mathbf{x}(t-\tau)$ decreases as $\tau$ increases beyond the model order $p$.  For large $\tau$, $\E [ \mathbf{x}(t) \mathbf{x}^T(t-\tau) ] \approx \mathbf{0}$, so an approximately independent set of samples can be found by collecting every $Cp^{th}$ sample from all nodes for some sufficiently large constant $C$.  In practice, it is more common to use every data point when estimating MAR models despite the dependency.

An additional challenge -- and motivation for group sparse approaches -- is the limited number of data samples available.  Specific connectivity patterns in a SMART model of a real network may change over time, which limits the number of samples for which the network is approximately stationary. Analysis of the performance of (\ref{eq:glasso_ar}) or (\ref{eq:glvar}) is difficult for limited data cases (finite $n$); however, the asymptotic theory and the simulations presented in Section~\ref{GSMAR:sims} suggest that when $\psi_{j \rightarrow i}^{FC}$ is small, connectivity estimation is easier.  Also, weak connections (for which $\| \mathbf{a}_{i,j} \|_2$ is small) are more difficult to recover with limited data.  For small enough $\lambda$ and large enough $n$, all connections will probably be recovered.  When $n$ is limited, the probability of recovering all connections, particularly weak ones, is decreased.

% Increasing $\lambda$ limits the number of connections identified by both the SG and SCSG approaches.  

Although Theorem~\ref{thm:ascon} indicates how $\lambda$ should scale with $n$, selecting $\lambda$ for non-asymptotic regimes can be difficult.  As seen in Section~\ref{GSMAR:sims}, $\lambda$ balances missed connections (Type II errors) with false positives (Type I errors).  Ideally, one would select $\lambda$ to achieve a specified famlywise error rate or false discovery rate; however, calculating p-values of each connection for a given $\lambda$ is an open problem.

Due to the difficulty of selecting an appropriate regularization parameter, it can be beneficial to consider the family of solutions achieved by varying $\lambda$.  The expectation-maximization (EM) algorithm described in \cite{akb_ni:09} efficiently solves the SG or, with slight modification, SCSG problem over a range of $\lambda$, successively adding connections as $\lambda$ decreases.  In that work, a heuristic is used to select a single $\lambda$ from the family of possible solutions \cite{akb_ni:09}.  In Sec.~\ref{GSMAR:sims} we use tenfold cross-validation to select the $\lambda$ which performs best on held out data.  Another possibility is to apply a Wald test for Granger-causality \cite{lutkepohl:91} successively to the last connection which enters the model and stop when a connection passes the test.  A recently proposed stability selection technique combines lasso and randomized subsampling to provide subset selection with false discovery rate bounds \cite{meins:10}.  This technique could potentially be applied to the SMART model at the expense of additional computation.

%Using the expectation-maximization (EM) algorithm described in \cite{akb_ni:09}, solutions can be found efficiently for a range of $\lambda$, and a family of possible networks can be considered.  In Sec.~\ref{GSMAR:sims} cross validation is used to select $\lambda$.  Alternatively, a heuristic such as that described in \cite{akb_ni:09} can be used.

\subsection{Normalization}
\label{GSMAR:normalization}

%It is common in practice to normalize measurements from each node to have equal power

Measurements from each node are often normalized to have equal power \cite{meins:06,pereda:05}.  We can account for normalization in any SMART model as follows.  Equal power in all channels means the diagonal of $\Gammab$ consists of all ones.  Thus we can transform $\Gammab$ to a normalized model using a diagonal matrix $\mathbf{D}^{-1/2}$ to obtain $\tilde{\Gammab} = \mathbf{D}^{-\frac{1}{2}} \Gammab \mathbf{D}^{-\frac{1}{2}}$.  Eq.~(\ref{eq:cov}) implies:

\begin{eqnarray*}
\tilde{\Gammab} &=& \mathbf{D}^{-\frac{1}{2}} \mathbf{A} \mathbf{D}^{\frac{1}{2}} \left( \mathbf{D}^{-\frac{1}{2}} \Gammab \mathbf{D}^{-\frac{1}{2}} \right) \mathbf{D}^{\frac{1}{2}} \mathbf{A}^T \mathbf{D}^{-\frac{1}{2}} \\
& & + \mathbf{D}^{-\frac{1}{2}} \widetilde{\mathbf{\Sigma}} \mathbf{D}^{-\frac{1}{2}} \\
 & = & \tilde{\mathbf{A}} \tilde{\Gammab} \tilde{\mathbf{A}}^T + \widetilde{\mathbf{\Sigma}}^*
\end{eqnarray*}

\noindent
where:

\begin{equation*}
\mathbf{D} = \begin{bmatrix} \mathbf{D}_1 & \mathbf{0} & \ldots & \mathbf{0} \\ \mathbf{0} & \mathbf{D}_2 & \ldots & \mathbf{0} \\ \vdots & \vdots & \ddots & \vdots \\ \mathbf{0} & \mathbf{0} & \ldots & \mathbf{D}_N \\ \end{bmatrix}
\end{equation*}

\noindent
Here $\mathbf{D}_i = \sigma_i^2 \mathbf{I}_p$ where $\sigma_i^2$ is the power in each node before normalization.

The effect of normalization on the ability of group sparse approaches to recover network structures is complicated.  We have found that normalization tends to decrease $\psi_{max} = \max_{(i,j) \in \A_{\mbox{\tiny active}}^C} \psi_{j \rightarrow i}$, indicating an improvement in asymptotic recoverability (for fixed $m$, $p$, and $N$ at least).  On the other hand, normalization clearly alters connection strength, meaning some connections may be weakened due to normalization and difficult to recover in the finite sample case.

\section{Example MAR networks}
\label{GSMAR:samplenets}

\begin{figure}[htbp]
\centering
\subfigure[ Circle Network ] {\label{fig:circlenet}
\includegraphics[height=2in]{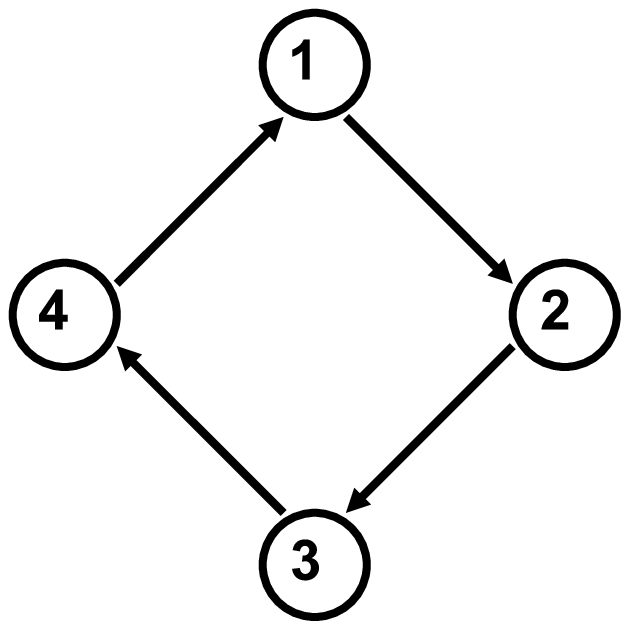}
} \qquad
\subfigure[ Parallel Network ] {\label{fig:parallelnet}
\includegraphics[height=2in]{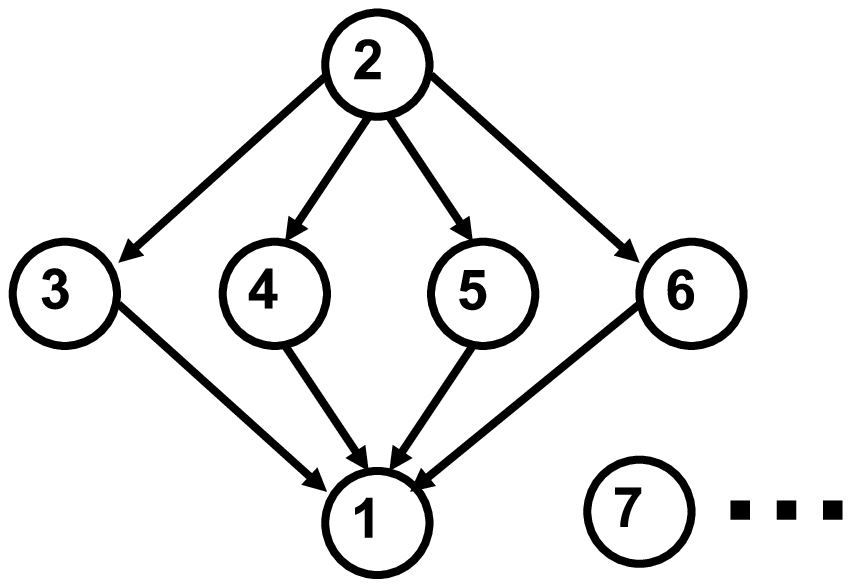}
}
\caption{Contrasting example MAR topologies, self-connections not shown.  }
\label{fig:circparnets}
\end{figure}

\begin{figure}[htbp]
\centering
\subfigure[ Winterhalder Network ] {\label{fig:winternet}
\includegraphics[width=0.45\columnwidth]{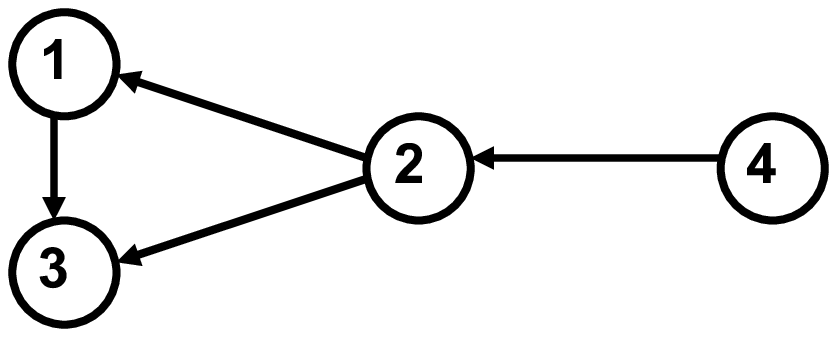}
}
\subfigure[ Haufe Network ] {\label{fig:haufenet}
\includegraphics[width=0.45\columnwidth]{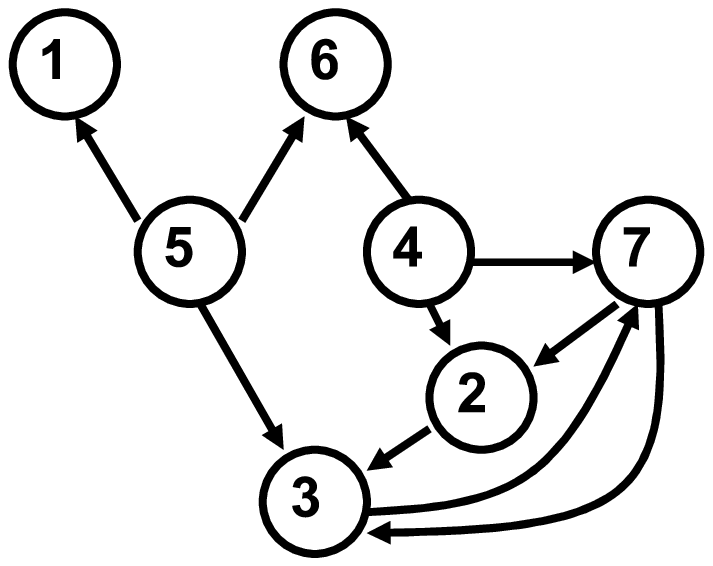}
}
%\subfigure[ Alternate Haufe Network ] {\label{fig:althaufenet}
%\includegraphics[width=0.45\columnwidth]{Figures/Haufe2.eps}
%}
\caption{MAR network topologies from existing literature.}
\label{fig:samplitnets}
\end{figure}

The false connection scores $ \psi_{j \rightarrow i}^{FC}$ and $\widetilde \psi_{j \rightarrow i}^{FC}$ are the key quantities that determine whether SG or SCSG will recover the connections which influence node $i$.  We consider four example networks in this section to develop insight on the nature of identifiable topologies.  Figure~\ref{fig:circparnets} depicts circular and parallel topologies constructed for this paper while Fig.~\ref{fig:samplitnets} depicts networks that have been studied in previous literature (see  \cite{winterhalder:05}, \cite{haufe:09}\footnote{In \cite{haufe:09} the direction of causal influence is unclear.  The network structure is described by a matrix of ones and zeros, but it is unclear whether a one in the $(i,j)^{th}$ position represents a connection from $i$ to $j$ or vice versa.  We show one possibility here and note that the other possible network (not shown) has similar properties.}). We compute the false connection scores for both the original network and after normalization (Sec.~\ref{GSMAR:normalization}) to determine whether the network is identifiable as $n \rightarrow \infty$ for SG and SCSG.  The maximum false connection scores for each network are listed in Table~\ref{maxscoretable}.

\begin{table}
\centering
\caption{Maximum false connection scores.}
\label{maxscoretable}
%\scriptsize
\vspace{\baselineskip}
\begin{tabular}{|c||c|c||c|c|}
\hline
\multirow{2}{*}{Network} & \multicolumn{2}{|c||}{Original} & \multicolumn{2}{|c|}{Normalized} \\
\cline{2-5} & \rule[-.2cm]{0cm}{.6cm}  $\psi_{max}^{FC}$ & $\tilde{\psi}_{max}^{FC}$ & $\psi_{max}^{FC}$ & $\tilde{\psi}_{max}^{FC}$ \\
\hline
Circle & 0.47 & 0.43 & 0.47 & 0.43 \\
\hline
Parallel & 1.93 & 1.06 & 1.04 & 1.03 \\
\hline
Winterhalder & 0.46 & 0.29 & 0.24 & 0.15 \\
\hline
Haufe & 0.83 & 0.56 & 0.71 & 0.57 \\
\hline
%Alt. Haufe & 0.73 & 0.65 & 0.73 & 0.65 \\
%\hline
\end{tabular}
\end{table}

Each node in the ``Circle Network'' shown in Fig.~\ref{fig:circlenet} is driven by it's own past as well as one other node forming the topology of a large feedback loop.  We chose MAR order $p=4$ and drew MAR coefficients from a normal distribution ($\mathcal{N}(\mathbf{0},0.04 \mathbf{I})$).  The first realization which resulted in a stable network is selected. The maximum false connection scores for this network are $\psi_{j \rightarrow i}^{FC}=0.47$ and  $\widetilde \psi_{j \rightarrow i}^{FC} = 0.43$.  Since these are less than one, the network connectivity can be recovered (as $n \rightarrow \infty$) using both SG and SCSG.

%(\ref{eq:glasso_ar})

The parallel network (Fig.~\ref{fig:parallelnet}) connectivity structure and coefficients were selected deliberately to confound group sparse approaches.   We chose $\mathbf{a}_{2 \rightarrow 2} = [\begin{smallmatrix} .2 & .2 & .2 & .2 \end{smallmatrix}]^T$ and $\mathbf{a}_{i \rightarrow i} = [\begin{smallmatrix} .05 & .05 & .05 & .05 \end{smallmatrix}]^T$ for $i \neq 2$.  All other connections shown are given by $\mathbf{a}_{i \rightarrow j} = [\begin{smallmatrix} .15 & .15 & .15 & .15 \end{smallmatrix}]^T$.   This network highlights several important aspects of SCSG, so we explore it in some detail.  The false connection scores for this network are summarized in Table~\ref{parnettable}.

\begin{table}
\centering
\caption{False connection scores for Parallel Network.}
\label{parnettable}
%\scriptsize
\vspace{\baselineskip}
\begin{tabular}{|c||c|c||c|c|}
\hline
\multirow{2}{*}{Connection} & \multicolumn{2}{|c||}{Original} & \multicolumn{2}{|c|}{Normalized} \\
\cline{2-5} & \rule[-.2cm]{0cm}{.6cm}  $\psi_{i \rightarrow j}^{FC}$ & $\tilde{\psi}_{i \rightarrow j}^{FC}$ & $\psi_{i \rightarrow j}^{FC}$ & $\tilde{\psi}_{i \rightarrow j}^{FC}$ \\
\hline
$1 \rightarrow 2$ & 1.41 & 0 & 0.74 & 0 \\
\hline
$2 \rightarrow 1$ & 1.06 & 1.06 & 1.04 & 1.03 \\
\hline
$1 \rightarrow 3$ & \multirow{4}{*}{1.93} & \multirow{4}{*}{0.71} & \multirow{4}{*}{1.00} & \multirow{4}{*}{0.37} \\
$1 \rightarrow 4$ & & & & \\
$1 \rightarrow 5$ & & & & \\
$1 \rightarrow 6$ & & & & \\
\hline
$3 \rightarrow 2$ & \multirow{4}{*}{0.61} & \multirow{4}{*}{0} & \multirow{4}{*}{0.63} & \multirow{4}{*}{0} \\
$4 \rightarrow 2$ & & & & \\
$5 \rightarrow 2$ & & & & \\
$6 \rightarrow 2$ & & & & \\
\hline
\end{tabular}
\end{table}
No matter which approach is used, a false connection from node~$2$ to node~$1$ will be established with high probability as $n \rightarrow \infty$.  This is due to the fact that there are four parallel paths connecting node~$2$ to node~$1$.  Since node~$2$ has such a strong combined influence on node~$1$, group sparse approaches are likely to identify a direct link.  False connections from node~$1$ to nodes~$3$--$6$ are also likely for large $n$ when SG is used.  On the other hand, the probability of linking $1$ to $3$--$6$ goes to zero as $n$ increases if SCSG is used.  This illustrates an important characteristic of SCSG: the asymptotic likelihood of false connections from a child to a parent tends to be reduced when self-connections are not penalized.  Proving this is always true seems difficult, but we provide some rationale.  The difference between $\psi_{j \rightarrow i}^{FC}$ and $\tilde{\psi}_{j \rightarrow i}^{FC}$ is the term $\Psib_{j,i}^T \frac{\mathbf{a}_{i,i}}{\| \mathbf{a}_{i,i} \|_2}$, whose norm lies between the singular values of the square matrix $\Psib_{j,i}$.  While it is difficult to verify that vector $\mathbf{a}_{i,i}$ lines up with a strong left singular vector of $\Psib_{j,i}$, we can expect that $\Psib_{j,i}$ will be ``large'' relative to other $\Psib_{j,k}$ since there is a connection from $i$ to $j$.

%(here $j$ is the parent node, $i$ is the child node), in which $\Psib_{j,i}$ indicates how to estimate the signal in parent node~$j$ given child node~$i$

The false connection score from node~$1$ to node~$2$ in Fig.~\ref{fig:parallelnet} highlights another important (and related) feature of SCSG.  The probability of falsely identifying connections to any node~$i$ which is only influenced by its own past goes to zero as $n$ goes to $\infty$ since $\tilde{\psi}_{j \rightarrow i}^{FC}$ is always zero.

The parallel network example also indicates that additional, unconnected nodes (i.e., node~$7$) do not change the false connection scores of connected nodes.  The chance of a false connection will increase in the finite $n$ case, but asymptotically such additional nodes do not matter since, as $n$ grows, the estimated correlation between two unconnected nodes will go to zero.

The network in Fig.~\ref{fig:winternet} (see \cite{winterhalder:05}) is not only group sparse, but sparse as well; every connection but one (self-connection of node $4$) consists of only one coefficient at one time lag, as shown by:

\begin{eqnarray*}
x_1(t) & = & 0.8 x_1(t-1) + 0.65 x_2(t-4) + u_1(t) \\
x_2(t) & = & 0.6 x_2(t-1) + 0.6 x_4(t-5) + u_2(t) \\
x_3(t) & = & 0.5 x_3(t-3) - 0.6 x_1(t-1) + 0.4 x_2(t-4) \\
& & + u_3(t) \\
x_4(t) & = & 1.2 x_4(t-1) - 0.7 x_4(t-2) + u_4(t) \\
\end{eqnarray*}

\noindent
As shown in Table~\ref{maxscoretable}, this network is recoverable by either method.

The structure of the network shown in Fig.~\ref{fig:haufenet} is taken from Fig.~1 of \cite{haufe:09}.  As in \cite{haufe:09}, we draw coefficients from a $\mathcal{N}(\mathbf{0},0.04 \mathbf{I})$ distribution and check for stability.  This network, which includes multiple paths of influence and feedback loops, can be recovered via both SG and SCSG with high probability as $n$ increases.

\section{Simulations}
\label{GSMAR:sims}

We now simulate the circle and parallel networks depicted in Fig.~\ref{fig:circparnets} to illustrate SG and SCSG network recovery performance with finite $n$.  (Simulations of the Haufe and Winterhalder networks performed similarly to the circle network and are omitted for space.)  Signals were simulated via (\ref{eq:mar}) with the initial condition for each simulation determined from the steady state distribution and with white driving noise of equal power in each node.  The expectation-maximization (EM) algorithm described in \cite{akb_ni:09} is used to solve the SG and SCSG optimization problems for $\lambda \in [0.05\lambda_{max}, \lambda_{max}]$, where $\lambda_{max}$ is the minimum $\lambda$ such that $\hat{\mathbf{a}}_i = \mathbf{0}$.  A specific $\lambda$ is selected separately for each node via tenfold cross validation using prediction error on held out data.  We assume the correct model order $p$ is known.  Thirty realizations of each network are generated with $n=150$ time samples.  We count the percentage of the 30 trials in which the true connections are correctly identified as well as the percentage of trials in which nonexistent connections are incorrectly identified.

%We also employ debiasing (see \cite{akb_ni:09}) to estimate connection coefficients once nonzero groups are identified.

\begin{figure}[htbp]
\centering
\subfigure[ Circle Network ] {\label{fig:circconprobSG}
\includegraphics[height=2in]{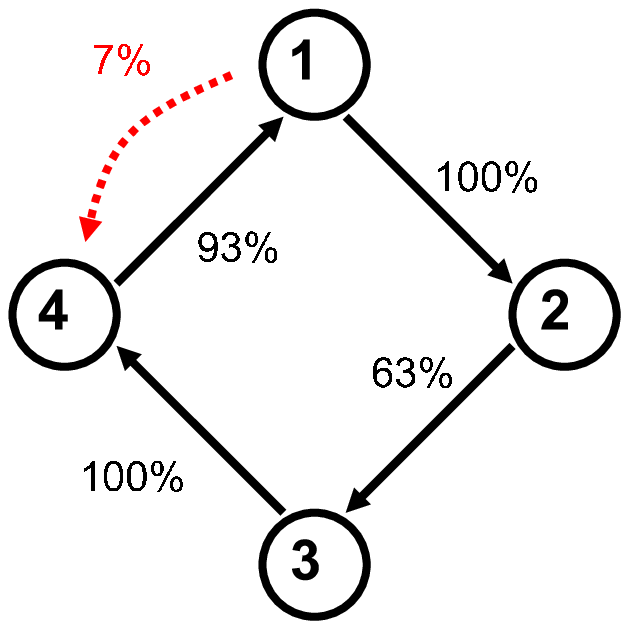}
} \qquad
\subfigure[ Circle Network ] {\label{fig:circconprob}
\includegraphics[height=2in]{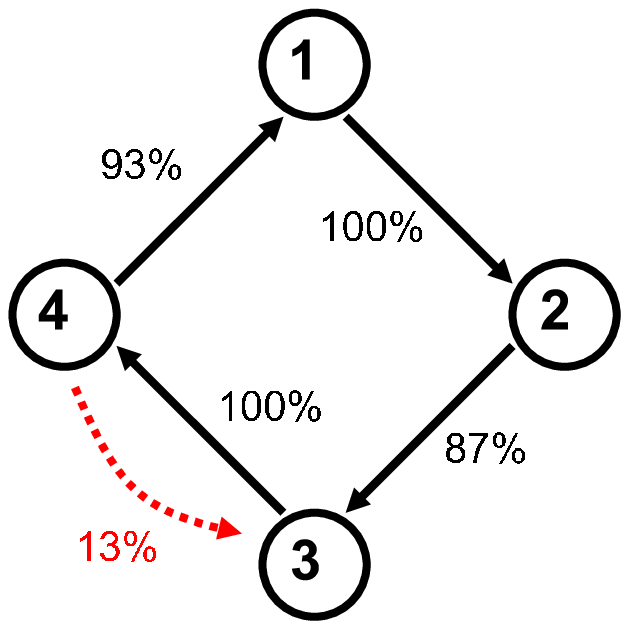}
}
\caption{Inferring the circle network using SG and SCSG with cross validation from $n=150$ time samples.  Black lines and numbers illustrate true connections and the percentage of 30 trials in which they are correctly identified.  Red dotted lines and text identify the most common false connection and percentage of occurrence over 30 trials.}
\label{fig:conprobsCIR}
\end{figure}

The results for SG and SCSG applied to the circle network are illustrated graphically in Fig.~\ref{fig:conprobsCIR}.  The true connections are identified in most of the cases for the circle network.  The strength of the four connections are given by $\| \mathbf{a}_{2,1} \|_2 = 0.46$, $\| \mathbf{a}_{3,2} \|_2 = 0.30$, $\| \mathbf{a}_{4,3} \|_2 = 0.37$, and $\| \mathbf{a}_{1,4} \|_2 = 0.28$.  The two true connections that are most often missed are the weakest connections of the four ($2 \rightarrow 3$ and $4 \rightarrow 1$).  The SCSG approach identifies the connection from $2 \rightarrow 3$ considerably more often, however.  The most common false connection with SG was from node $1$ to node $4$ and occurred in only 2 of 30 trials, while a false connection from node 4 to node 3 was identified in 4 of 30 trials using SCSG.  Qualitatively similar results are obtained for $n=50$ and $n=100$ with the performance improving for most connections as the number of samples increases.  A noticeable improvement in ability to identify true connections results as the number of samples increases from $n=50$ to $n=150$.

%Inferring circle and parallel networks using SG with cross validation from $n=150$ time samples.  Black lines and numbers illustrate true connections and the  the percentage of 30 trials in which they are correctly identified.  Red dotted lines and text identify the most common false connection and percentage of occurrence over 30 trials.

%we generate 90 realizations of each network: 30 realizations with $n=50$ data samples, 30 with $n=100$, and 30 with $n=150$.  In Fig.~\ref{fig:conprobs} we illustrate the percentage of trials for which (\ref{eq:glvar}) with cross validation identifies the correct connections, as well as the most common false connection for the $n=150$ case.

%For the $n=50$ and $n=100$ cases, qualitatively similar results were seen with the performance improving for most connections as the number of samples increased.

\begin{figure}[htbp]
\centering
\subfigure[ Parallel Network ] {\label{fig:parconprobSG}
\includegraphics[height=2in]{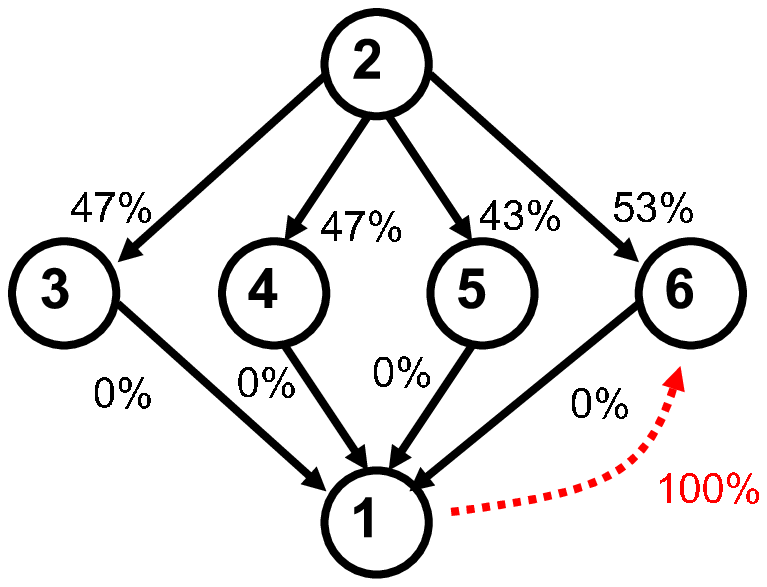}
} \qquad
\subfigure[ Parallel Network ] {\label{fig:parconprob}
\includegraphics[height=2in]{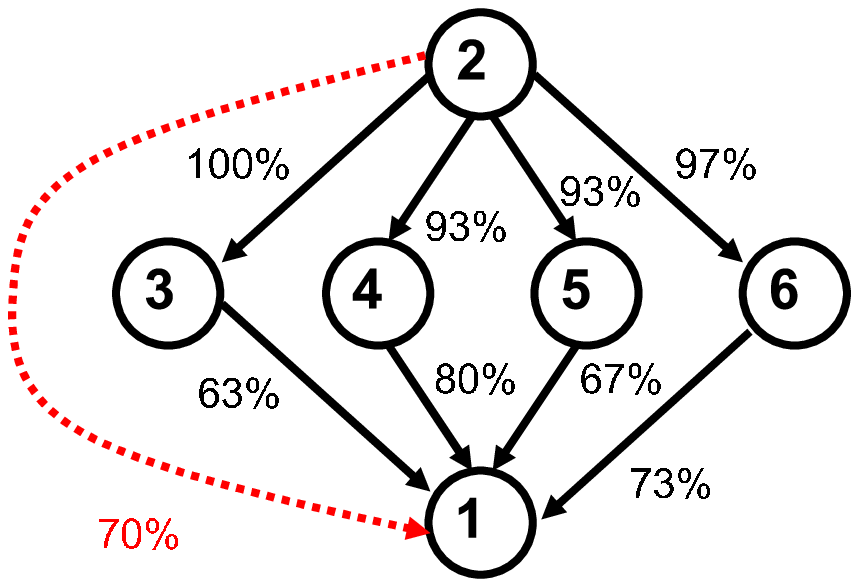}
}
\caption{Inferring the parallel network using SC and SCSG with cross validation from $n=150$ time samples.  Black lines and numbers illustrate true connections and the  the percentage of 30 trials in which they are correctly identified.  Red dotted lines and text identify the most common false connection and percentage of occurrence over 30 trials.}
\label{fig:conprobsPAR}
\end{figure}

%Inferring circle and parallel networks using SCSG with cross validation from $n=150$ time samples.  Black lines and numbers illustrate true connections and the  the percentage of 30 trials in which they are correctly identified.  Red dotted lines and text identify the most common false connection and percentage of occurrence over 30 trials.

%For the $n=50$ and $n=100$ cases, false connection rates similar to the $n=150$ case were obtained.  There was a noticeable improvement in the ability to identify the true connections as the number of samples increased from $n=50$ to $n=150$.

As predicted by the theoretical arguments of Sec.~\ref{GSMAR:recfinite}, the SG approach does not perform as well on the parallel network (Fig.~\ref{fig:conprobsPAR}).  In particular, the true connections from nodes 3, 4, 5, and 6 to node 1 are never identified, the true connections from node 2 to nodes 3, 4, 5, and 6 are identified about half of the time, and  the connection from node 1 to 6 is incorrectly identified in all cases.  The next most common false connections (not shown in Fig.~\ref{fig:conprobsPAR}) are from node~$1$ to nodes~$3$--$5$ with probabilities of 93\%, 83\%, and 87\%, respectively.  These four false connections (from node~$1$ to its parents) have the highest false connection score ($\psi^{FC}_{1 \rightarrow j} = 1.93$, $j=3,4,5,6$) for this scenario, according to Table~\ref{parnettable}.  The false connection from node~$1$ to node~$2$ is the next most common, occurring in 80\% of the trials.  The false connection score for this link is 1.41.  Notice these five most common false connections reverse the true direction of causal influence.

The SCSG approach performs considerably better for the parallel network, consistent with the improvement in the false connection scores given in  Table~\ref{parnettable}.  The connections from node~$2$ to nodes~$3$--$6$ are almost always discovered, although the true connections from nodes~$3$--$6$ to node~$1$ are missed more frequently.  However, SCSG identifies a connection directly from node~$2$ to node~$1$ in 70\% of the trials.  A possible explanation for this error is that a single connection from node~$2$ to node~$1$ is a sparser solution than connecting nodes $3$--$6$ to node~$1$ and accounts for much of the variance at node~$1$.  The connection from node~$2$ to node~$1$ has the highest false connection score (see Table~\ref{parnettable}).

When using SG on the parallel network, none of the true connections to node~$1$ are identified.  While these connections might be recovered by allowing a greater range of $\lambda$ in the cross validation selection procedure, their absence reveals a downside to penalizing self-connections.  As $\lambda$ is decreased below $\lambda^*$, the first connection identified is the self-connection.  When SCSG is used, self-connections are always present, so decreasing $\lambda$ below $\lambda^*$ activates a connection from a different node.  In a sense, the SCSG approach has a ``head start'' in detecting connections.

Simulations with $n=50$ and $n=100$ time samples (not shown) reveal that the ability of SCSG to recover the true connections improves as the number of samples increases.  However, the number of trials in which false connections were made between nodes~$1$ and $2$ (both directions) also increases as the number of samples increases.  This behavior is consistent with the asymptotic result of Cor.~\ref{cor:fixednet} which indicates that the probability of identifying the wrong network goes to one as the number of samples increases.

%Frequency of true connections and most common false connection identified for $n=150$ samples using gLasso (\ref{eq:glasso_ar}).

\section{Macaque Brain Simulation}
\label{macaque}

%(available online: http://sites.google.com/a/brain-connectivity-toolbox.net/bct/datasets)

%Recently, Lasso-type procedures have been applied to MAR model estimation of brain activity.  Valdes-Sosa et al. \cite{valdes:05} apply the Lasso, as well as other sparsity inducing nonconvex techniques, to simulated and fMRI data.  The Elastic Net, essentially the Lasso plus an additional $\ell_2$ penalty, is applied to fMRI data in \cite{carroll:09}.  In 2008, Haufe, et al. \cite{haufe:09} proposed a gLasso variant for MAR network estimation and compared it to other techniques via simulation of relatively small networks, some of which we reproduce below.  They employ a gLasso variant as a regularization term in a more complicated cost function in \cite{haufe:09b}.  None of these works explore theoretical behavior of sparsity inducing techniques.

\begin{figure*}
\centering
\subfigure[ Full Network ] {\label{fig:macaquenet}
\includegraphics[height=2.5in]{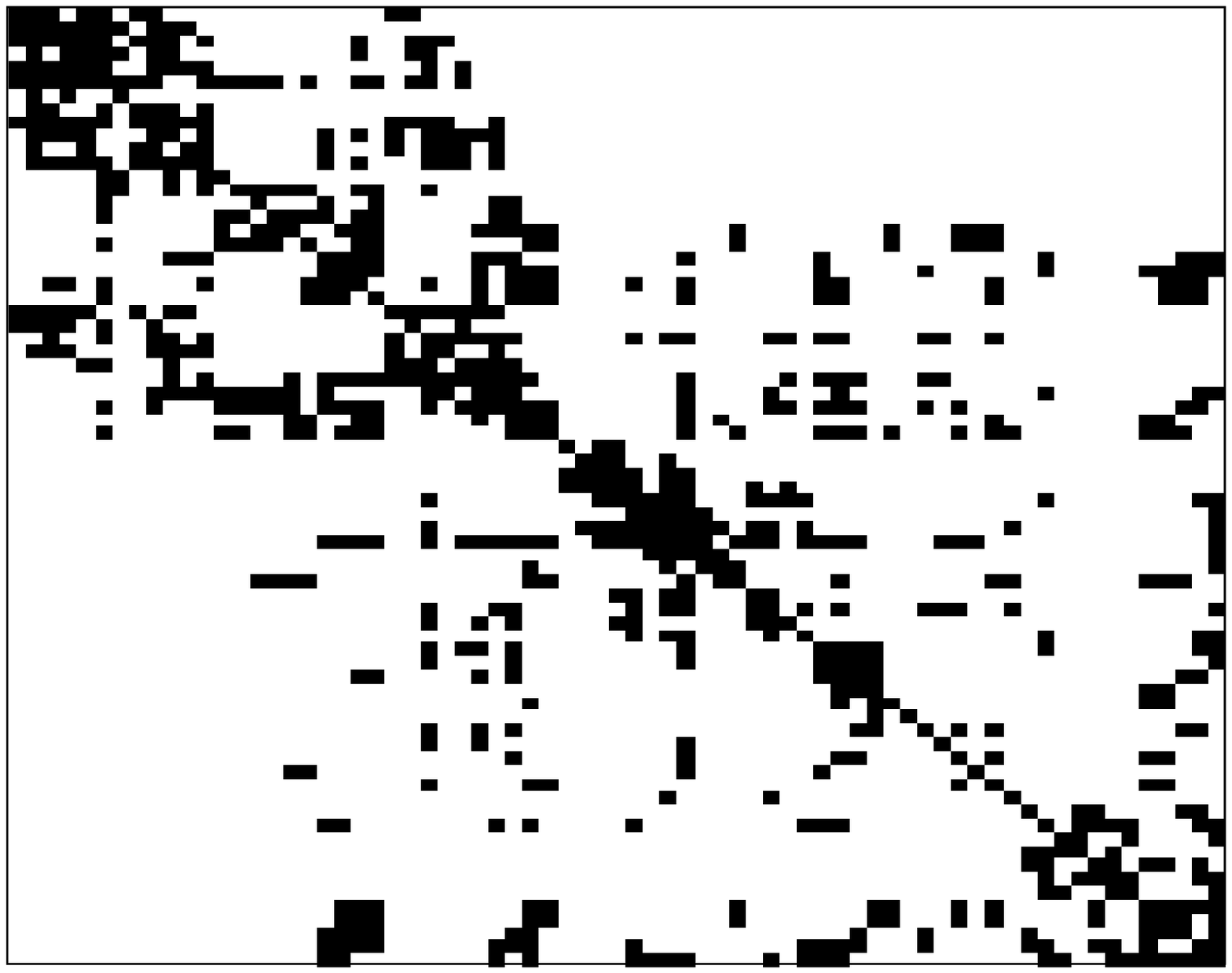}
}
\subfigure[ Subset Network ] {\label{fig:macaqueSubnet}
\includegraphics[height=2.5in]{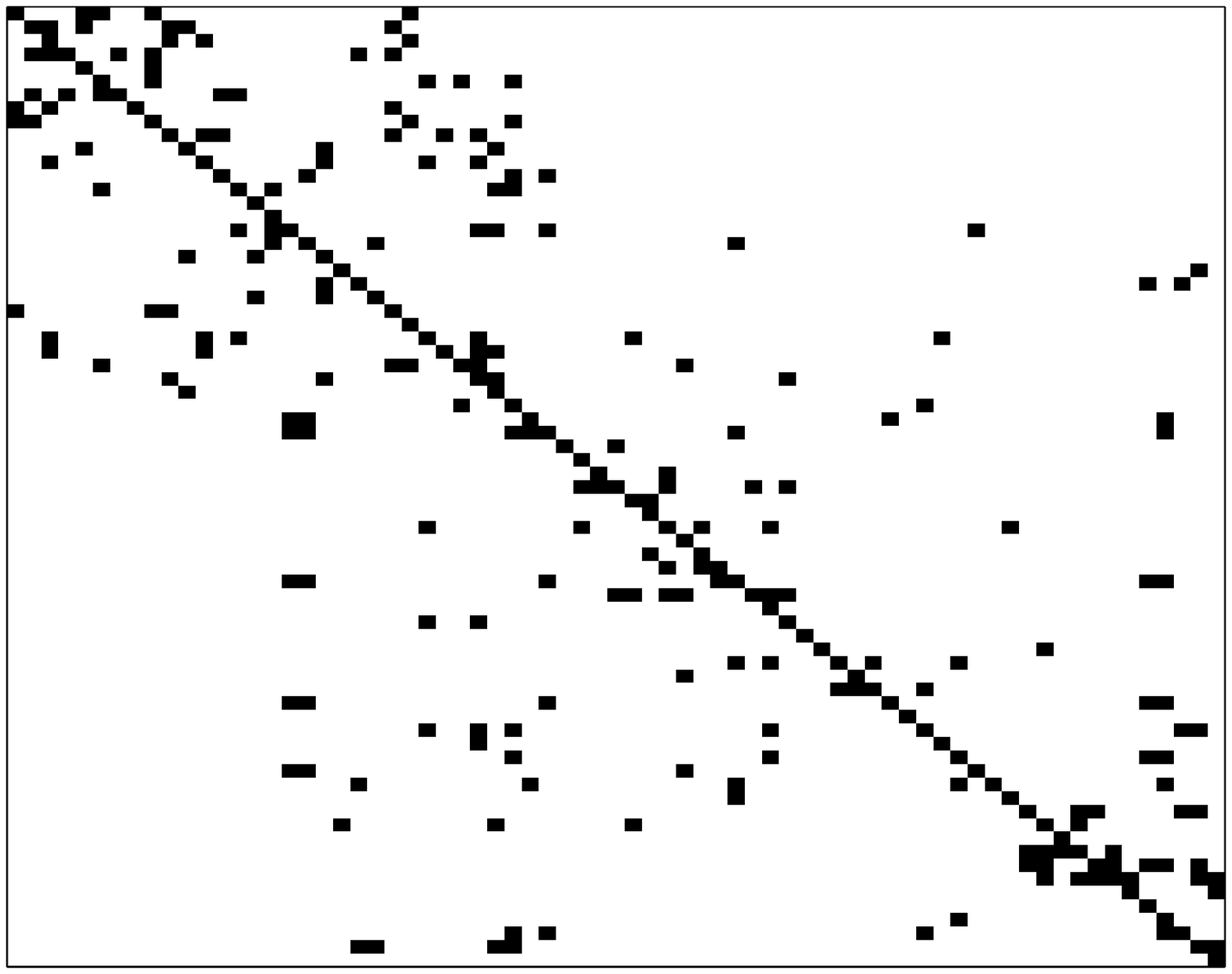}
}
\caption{Connectivity matrices of the simulated macaque brain networks: (a) all physical connections are active and (b) up to ten parent nodes are active.  A connection from node $i$ to node $j$ exists if the entry in the $i^{th}$ row and $j^{th}$ column is black.}
\end{figure*}

Lasso-type procedures have recently been applied to MAR model estimation of brain activity \cite{valdes:05,carroll:09,haufe:09,haufe:09b}.  In this section we simulate electrocorticogram (ECoG) recordings with a SMART model using a realistic network topology obtained from tract-tracing studies of a macaque brain \cite{young:93,sporns:02}.  A matrix representing connectivity in the macaque brain -- the ``macaque71'' data set, consisting of 71 nodes and 746 connections -- is shown in Fig.~\ref{fig:macaquenet}.  Each node is an area of the cortex.  A connection between areas exists if neuronal axons physically connect respective areas.  Figure~\ref{fig:macaquenet} suggests a sparse connectivity structure in the macaque.   Including self-connections, there are an average of 11.5 out of 71 possible parents for each node.

We simulate two networks based on this physical connectivity structure.  First we assume that every physical connection in the macaque71 data set is actively conveying information.  It is unrealistic to model every physical connection as active at a given time, so we also simulate a model in which up to ten randomly selected parents (including the self-connection) are active for each node.  For simulation purposes, we choose a model order of six and draw coefficients for nonzero entries of the $\mathbf{A}_i$ matrices independently from a $\mathcal{N}(\mathbf{0},0.04\mathbf{I})$ distribution for the full model and a $\mathcal{N}(\mathbf{0},0.16\mathbf{I})$ model for the subset model. The first realization for each model that results in a stable network is used.

Given these stable SMART models based on physical connections in the macaque brain, we generate time series using Eq.~\ref{eq:mar} with initial conditions $x_i(0) = 0$ and driving noise $u_i(t)$ distributed i.i.d. $\mathcal{N}(0,1)$ over all channels and all time samples.  The data are normalized, as described in Sec.~\ref{GSMAR:normalization} using the estimated power at each node.

Normalization reduces the worst case SCSG false connection score of the full network from 1.73 to 1.25.  Hence the SCSG estimate will be inconsistent as the number of samples increases.  Note however, that SCSG can still consistently recover the parents of nodes $i$ for which $\psi_{j \rightarrow i}^{FC} < 1$ for all $j \in \A_i^C$.  In this example, only four nodes $i$ have $\psi_{j \rightarrow i}^{FC} > 1$, meaning that the parents of 67 of the nodes can be recovered accurately.  Interestingly the neighborhoods of the four nodes which violate the false connection score condition exhibit a topology very similar to the parallel network described in Sec.~\ref{GSMAR:samplenets}.  Each of these four nodes has many parent nodes which provide an indirect link to the same ``grandparent'' node.  If only some of these paths are active at a given time, the network may be recoverable.  This is indeed the case in the subset model where the false connection score is reduced from 4.07 to 0.54 by normalization.

%thus the parents of all nodes can be consistently recovered from the normalized time series.

%, while the worst case score for the subset network is XX before and YY after normalization

%In the SMART gLasso estimate of the parents of nodes $i$ for which $\psi_i^{FC} > 1$, at least one node will be misidentified as a parent

%mac300.eps

% change to height=2.5 for two col., width=3.1 for one col
\begin{figure*}[htbp]
\centering
\subfigure[ Full Network, $T=300$ ] {\label{fig:macroc300}
\includegraphics[width=3.1in]{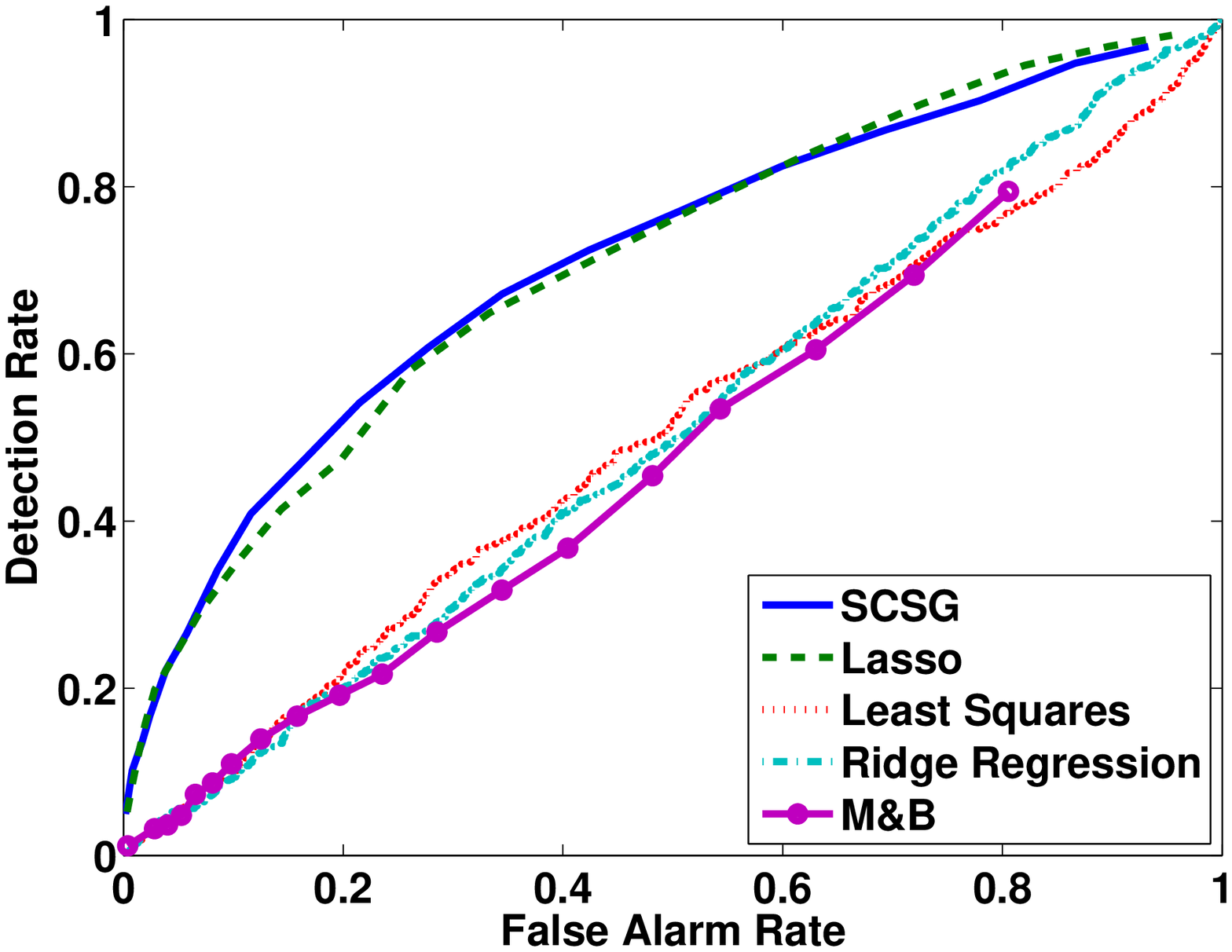}
}
\subfigure[ Full Network, $T=900$ ] {\label{fig:macroc900}
\includegraphics[width=3.1in]{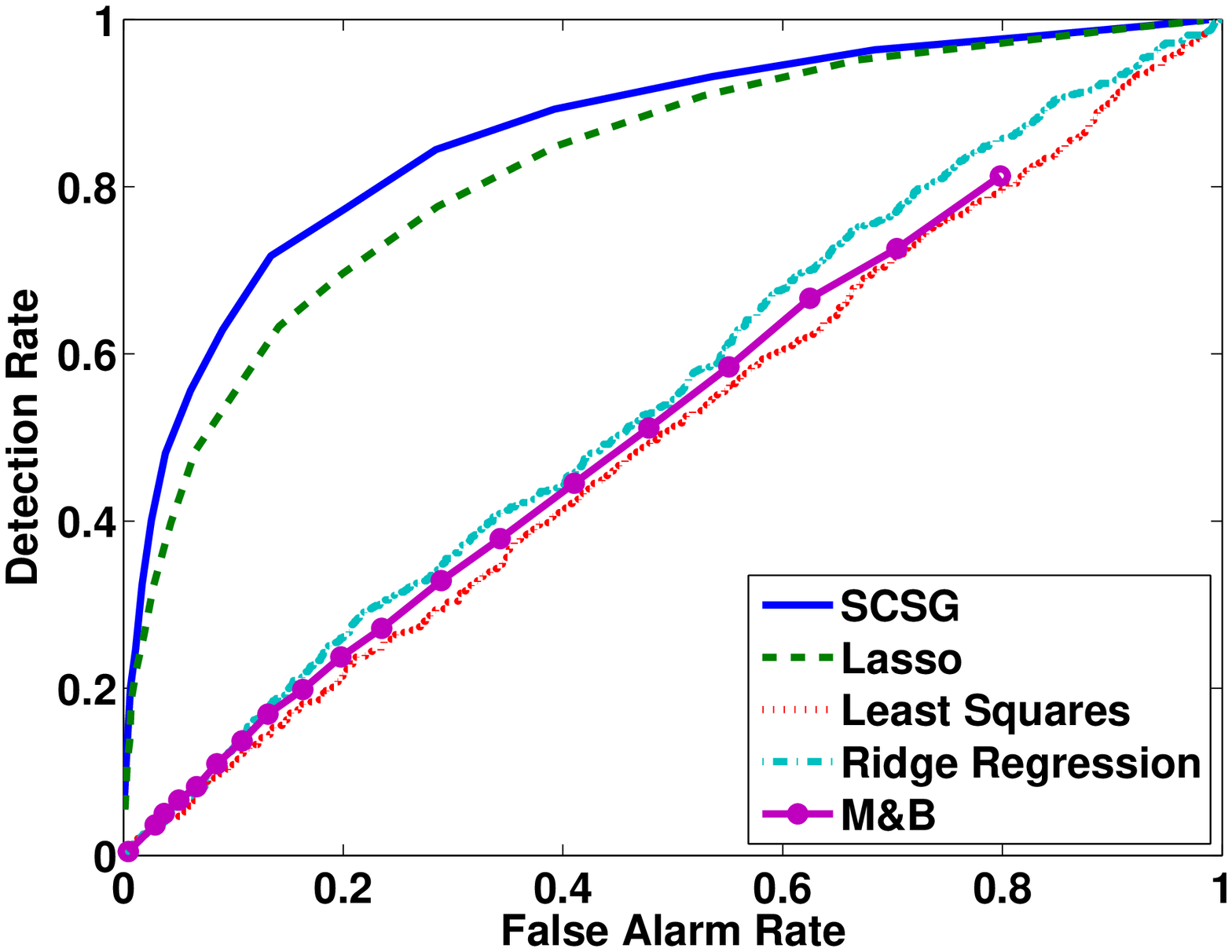}
}

\subfigure[ Subset Network, $T=300$ ] {\label{fig:macsubroc300}
\includegraphics[width=3.1in]{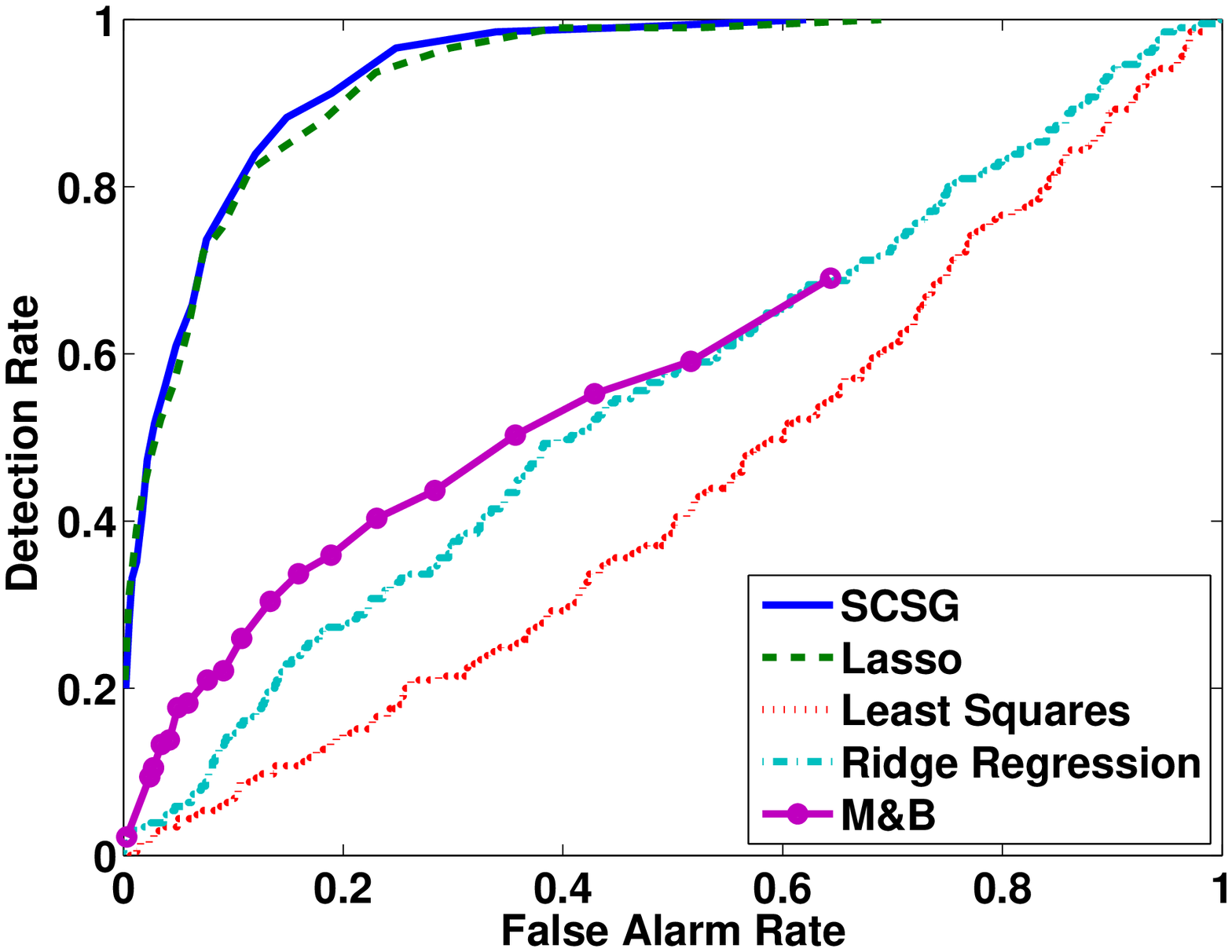}
}
\subfigure[ Subset Network, $T=900$ ] {\label{fig:macsubroc900}
\includegraphics[width=3.1in]{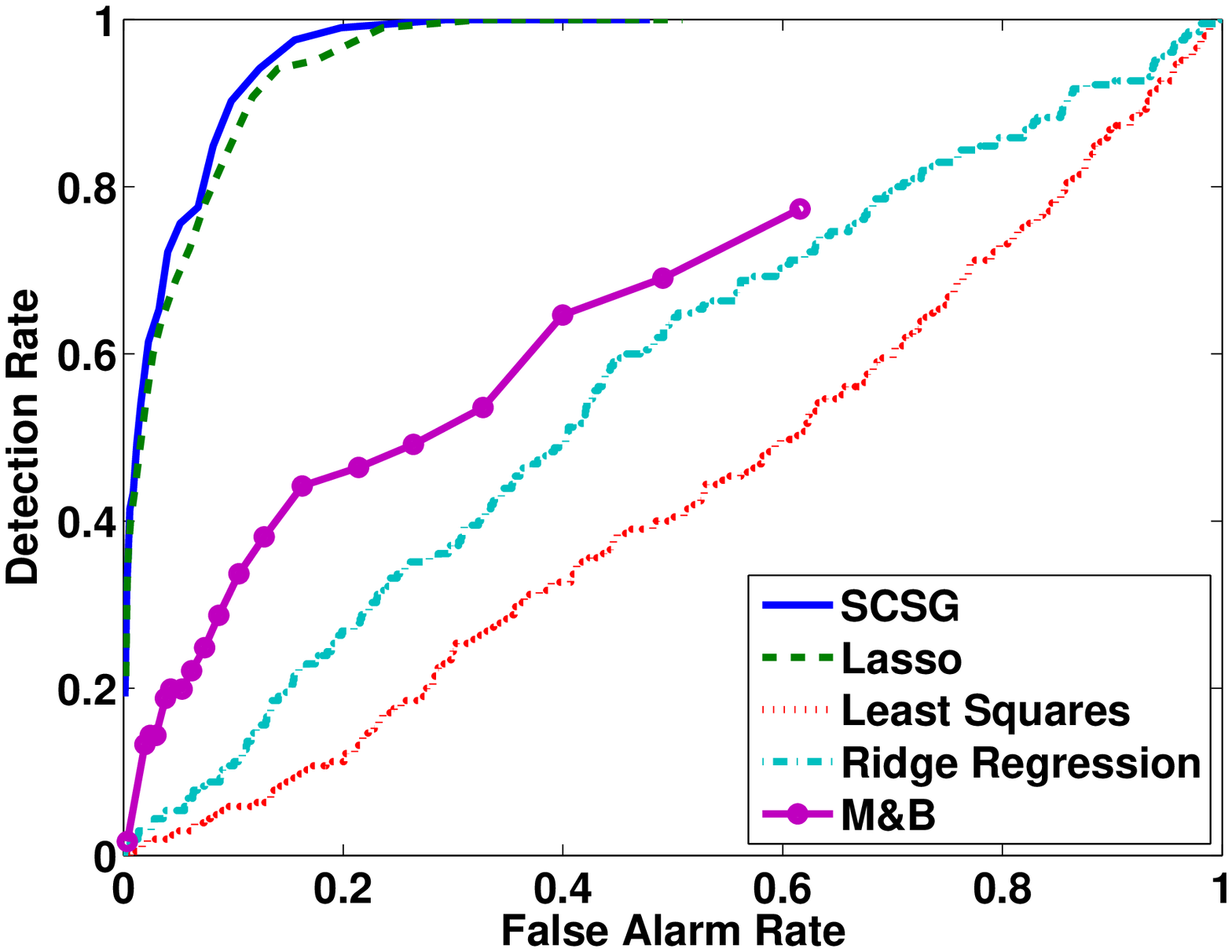}
}
\caption{Fraction of connections identified vs. fraction of zero valued $\mathbf{a}_{i,j}$ misidentified as nonzero (ROC curve) in simulated macaque brain networks.  Top row: all physical connections active.  Bottom row: subset of connections active.}
\label{fig:rocs}
\end{figure*}

%\setcaptionwidth{3in}

%\begin{figure}
%\centering
%\includegraphics[height=2.5in]{Figures/mac300_RQnew.eps}
%\caption{Fraction of connections identified vs. the fraction of zero valued $\mathbf{a}_{i,j}$ misidentified as nonzero (ROC) in a simulation of 300 time samples.}
%\label{fig:macroc300}
%\end{figure}

We illustrate the performance of several network estimation techniques in Fig.~\ref{fig:rocs} using receiver operating characteristic (ROC) curves.  We simulate the SCSG approach, the standard Lasso which promotes sparse coefficients as opposed to sparse connections (see Sec.~\ref{GSMAR:samplenets}), least squares estimation (Yule-Walker equations for $n > p N$), ridge regression, and an approach for estimating sparse non-causal networks described in \cite{meins:06} which we call the Meinshausen and B\"{u}hlmann (M\&B) approach.  The poor performance of the M\&B approach illustrates that it is not appropriate for causal network inference\footnote{Readers familiar with \cite{meins:06} will observe that the M\&B technique is not meant to recover nonzero connections as defined here, but rather nonzero entries in the inverse covariance matrix.  We point out that although the MAR networks presented here are sparse in the number of parent nodes, the inverse covariance matrices are not sparse.}.  The performance of the SG technique is similar to that of the SCSG for these networks, so we do not include it here.

%The Meinshausen and B\"{u}hlmann (M\&B) approach\footnote{Readers especially familiar with \cite{meins:06} will observe that the M\&B technique is not meant to recover nonzero connections as defined here, but rather nonzero entries in the inverse covariance matrix.  We point out that although the network is sparse, the inverse covariance matrix in this case is not sparse.} is essentially the SCSG with $p=1$.  

Using ROC curves to evaluate performance removes the difficult task of selecting regularization parameters (which relate, sometimes directly, to significance level) for different techniques.  The ROC curve is obtained for the SCSG, Lasso, and M\&B approaches by varying the penalty weight $\lambda$ (using the same solver with group size of one when necessary).  A detection occurs when a nonzero estimate $\hat{\mathbf{a}}_{i,j}$ coincides with a true connection from node $j$ to node $i$, while a miss occurs when $\hat{\mathbf{a}}_{i,j} = \mathbf{0}$ despite a true connection from $j$ to $i$.  False postives and true negatives are similarly defined.  For least squares and ridge regression approaches, we use the simultaneous inference method proposed in \cite{haufe:09} which makes use of adjusted p-values \cite{hothorn:08}; however, we threshold the normalized test statistics directly (rather than the p-values) to produce ROC curves in order to avoid compuationally intensive Monte Carlo sampling of multivariate integrals.  This yields the same curve due to the monotonic ralationship between test statistic and associated p-value.  Since SCSG has additional knowledge that all self-connections are non-zero, we do not include self-connections when calculating ROC curves for any method.  The ROC is defined as the percentage of true connections detected versus the percentage of false positive connections.

We simulate both $n=300$ and $n=900$ time samples from all 71 nodes.  In the first case we have fewer samples ($300 \times 71 = 21300$) than coefficients ($6 \times 71^2 = 30246$), so enforcing a sparse solution is essential.  This is clearly seen in Figs.~\ref{fig:macroc300} and \ref{fig:macsubroc300} where SCSG and Lasso clearly outperform the other methods.  In fact, least squares, ridge regression, and the Meinshausen and B\"{u}hlmann approach perform similarly to coin flipping.  The SCSG performs better than the Lasso because the group assumption of the gLasso better matches the true model.  In the second case with $n=900$ time samples for each node, we have a few more than two samples for every coefficient.  The results are shown in Figs.~\ref{fig:macroc900} and \ref{fig:macsubroc900}.  Both SCSG and Lasso perform better with more samples, as expected.  The other methods still perform similarly to coin flipping.  In the case of least squares and ridge regression, there are still too few samples to reliably estimate the covariance matrices.

%\begin{figure}
%\centering
%\includegraphics[height=2.5in]{Figures/mac300_active.eps}
%\caption{Fraction of connections identified vs. the fraction of zero valued $\mathbf{a}_{i,j}$ misidentified as nonzero (ROC) in the subset network in a simulation of 300 time samples.}
%\label{fig:macsubroc300}
%\end{figure}

\section{Conclusion}

We have analyzed application of the Group Lasso to the SMART model and proposed a modified gLasso for SMART model estimation.  The gLasso groups together all $p$ coefficients which comprise a connection from one node to another and penalizes the sum of the $\ell_2$ norm of these coefficient groups.  Such an approach tends to yield estimated networks with only a few nonzero connections.  Our proposed SCSG removes the penalty for self-connections so that  a node's own past is always used to predict its next state.  We have shown that both the SG and SCSG approaches are capable of recovering the true network structure under certain conditions, the most crucial of which we term the false connection score, $\psi_{max}$.  MAR networks are identifiable when $\psi_{max} < 1$, but not when $\psi_{max} > 1$.  To our knowledge, this is the first attempt to quantify the characteristics of MAR networks that result in gLasso based recovery.

The false connection score condition (and to some degree Assumption~\ref{as:minsv}) implies that the network under study must be not only sparse, but also have the property that each node in the network is independent enough from other nodes (then $\Psib_{i,j}$ will be small).  Clearly, a network with only self-connections satisfies this condition, but these are not very interesting or realistic.  On the other hand, small world networks \cite{watts:98} have the type of structure that seems likely to meet the false connection condition (again depending on the connection coefficients).  In small world networks, each node is connected to most of its nearest neighbors, but also has a few long range connections (short path lengths).  It has been shown that such networks efficiently transmit information to all nodes \cite{watts:98,sporns:07} and suggested that the brain may have a small-world network structure.  In fact, the structural connectivity pattern of the macaque brain used for simulations in Sec.~\ref{GSMAR:sims} represents a small-world network \cite{sporns:00,sporns:02}.  Small-world networks have sparse structure, though each node may have a somewhat large number of local connections.

The false connection score indicates whether a false positive connection is likely to occur.  False negatives or missed connections are also of concern.  Our analysis shows that, for fixed parameter networks ($m$, $p$, and $N$ constant), the penalty weight $\lambda$ can be set small enough that false negatives are improbable.  The false connection score determines whether this small $\lambda$ will avoid false positives.  Our experience suggests that misses are more likely to occur for weak connections.  Our examples indicate that the SCSG approach is effective at recovering network structure and that the false connection score is a an informative indicator of recovery performance for even relatively small sample sizes $n$.  Finally, note that the result of Theorems~\ref{thm:ascon} and \ref{thm:nec} apply to any gLasso application which satisfy the assumptions.  In a generic application the false connection score may be interpreted as a statistical property of the $\mathbf{X}$ matrix.

\appendices
\section{Proof of asymptotic consistency}
\label{GSMAR:asymptcons}

To prove Theorem~\ref{thm:ascon}, we consider applying gLasso (\ref{eq:glasso_ar}) to a single node (without loss of generality, node~$1$), and use the union bound to achieve the desired result.  We restate Assumption~\ref{as:scaling} in terms of positive constants $c_1$ -- $c_4$ to facilitate the proof: number of nodes $N = \mathcal{O}(n^{c_1})$, maximum number of parent nodes $m = \mathcal{O}(n^{c_2})$, model order $p = \mathcal{O}(n^{c_3})$, and regularization parameter $\lambda = \Theta(n^{-c_4/2})$ with $c_2 < c_4$ and $c_3 + c_4 < 1$.

\def\thesubsection{\Roman{subsection}}

\subsection*{KKT Conditions}

The Karush-Kuhn-Tucker (KKT) conditions for a solution to (\ref{eq:glasso_ar}) follow from the theory of subgradients.  The subgradient of $\| \mathbf{v} \|_2$ is any vector whose $\ell_2$-norm is less than one for $\mathbf{v} = \mathbf{0}$, while it is simply the gradient $\frac{\mathbf{v}}{\| \mathbf{v} \|_2}$ when $\mathbf{v} \neq \mathbf{0}$.  Thus the KKT conditions are given by:

\begin{eqnarray}
\mathbf{X}_i^T (\mathbf{y}_1 - \mathbf{X} \hat{\mathbf{a}}_1) = \frac{\lambda n \hat{\mathbf{a}}_{1,i}}{2 \| \hat{\mathbf{a}}_{1,i} \|_2} & & \forall \ i \: \mathrm{s.t.} \: \hat{\mathbf{a}}_{1,i} \neq \mathbf{0} \label{eq:inclcond} \\
\| \mathbf{X}_i^T (\mathbf{y}_1 - \mathbf{X} \hat{\mathbf{a}}_{1}) \|_2 \leq \frac{\lambda n}{2} & & \forall \ i \: \mathrm{s.t.} \: \hat{\mathbf{a}}_{1,i} = \mathbf{0}. \label{eq:exclcond}
\end{eqnarray}

\noindent
For convenience, we define $\hat{\mathbf{z}}_1=[\begin{smallmatrix} \hat{\mathbf{z}}_{1,1}^T & \ldots & \hat{\mathbf{z}}_{1,N}^T \end{smallmatrix}]^T$ with $\hat{\mathbf{z}}_{1,i} = \frac{2}{\lambda n} \mathbf{X}_i^T (\mathbf{y}_1 - \mathbf{X} \hat{\mathbf{a}}_1)$.  The vector $\hat{\mathbf{z}}_1$ restricted to the active set is denoted $\hat{\mathbf{z}}_{\mathcal{S}_1}$.  We assume without loss of generality that ${\mathcal{S}_1} = \{ 1,2,\ldots,m \}$.

\subsection*{Limiting False Negatives}

We start with conditions assuring that all nonzero coefficients are estimated as nonzero.  To do so, we follow the arguments used by \cite{obozinski:08}.  We consider the ``oracle'' solution; e.g., we consider the solution to the group sparse penalized estimator if the active set were known:

\begin{eqnarray}
\hat{\mathbf{a}}_1^{*}(\lambda) & = & \mathrm{arg} \min_{\alphab : \alphab_{\mathcal{S}_1^C}=0} \frac{1}{n} \left\| \mathbf{y}_1 - \mathbf{X} \alphab_1 \right\|^2 + \frac{\lambda}{2} \sum_{i=1}^N \| \alphab_{1,i} \|_2 \nonumber \\
& = & \mathrm{arg} \min_{\alphab_{S_1}} \frac{1}{n} \left\| \mathbf{y}_1 - \begin{bmatrix} \mathbf{X}_{\mathcal{S}_1} & \mathbf{X}_{\mathcal{S}_1^C} \end{bmatrix} \begin{bmatrix} \alphab_{\mathcal{S}_1} \\ \mathbf{0} \end{bmatrix} \right\|^2 \\ \nonumber
& + & \frac{\lambda}{2} \sum_{i \in \mathcal{S}_1} \| \alphab_{1,i} \|_2. \nonumber \\
\end{eqnarray}

\noindent
We must ensure that all coefficient subvectors in $\mathcal{S}_1$ are nonzero in the oracle estimate $\hat{\mathbf{a}}_1^{*}$.  Since all subvectors $\hat{\mathbf{a}}_{1,i}^{*}$ of $\hat{\mathbf{a}}_1^{*}$ will be zero for large enough $\lambda$, this means we must make sure that $\lambda$ is not too big.

All nonzero blocks must satisfy ($\ref{eq:inclcond}$), so we consider:

\begin{eqnarray}
%\label{eq:inclcond}
\frac{\lambda n}{2} \hat{\mathbf{z}}_{\mathcal{S}_1} & = & \mathbf{X}_{\mathcal{S}_1}^T (\mathbf{y} - \mathbf{X} \hat{\mathbf{a}}^{*}) \nonumber \\
& = & \mathbf{X}_{\mathcal{S}_1}^T (\mathbf{X}_{\mathcal{S}_1} \mathbf{a}_{\mathcal{S}_1} + \mathbf{u}_1 - \mathbf{X}_{\mathcal{S}_1} \hat{\mathbf{a}}_{\mathcal{S}_1}^{*}) \nonumber \\
& = & (\mathbf{X}_{\mathcal{S}_1}^T \mathbf{X}_{\mathcal{S}_1} (\mathbf{a}_{\mathcal{S}_1} - \hat{\mathbf{a}}_{\mathcal{S}_1}^{*}) + \mathbf{X}_{\mathcal{S}_1}^T \mathbf{u}_1
\label{eq:oracleincl}
\end{eqnarray}

\noindent
from which we obtain:

\begin{equation}
\label{eq:ahatoracle}
\hat{\mathbf{a}}_{\mathcal{S}_1}^{*} = \mathbf{a}_{\mathcal{S}_1} - \frac{\lambda n}{2} (\mathbf{X}_{\mathcal{S}_1}^T \mathbf{X}_{\mathcal{S}_1})^{-1} \hat{\mathbf{z}}_{\mathcal{S}_1} + (\mathbf{X}_{\mathcal{S}_1}^T \mathbf{X}_{\mathcal{S}_1})^{-1} \mathbf{X}_{\mathcal{S}_1}^T \mathbf{u}_1
\end{equation}

\noindent
where the invertibility of $\mathbf{X}_{\mathcal{S}_1}^T \mathbf{X}_{\mathcal{S}_1}$ is assured for large $n$ since $n$ grows faster than $mp$ by Assumption~\ref{as:scaling}.  At this point the following notation is convenient.  Let $\widehat{\mathbf{G}}_{\mathcal{S}_1} = n (\mathbf{X}_{\mathcal{S}_1}^T \mathbf{X}_{\mathcal{S}_1})^{-1}$, with columns partitioned as $\widehat{\mathbf{G}}_{\mathcal{S}_1} = [\begin{smallmatrix} \widehat{\mathbf{G}}_{\mathcal{S}_{1,1}} & \ldots & \widehat{\mathbf{G}}_{\mathcal{S}_{1,m}} \end{smallmatrix}]$, where each sub-matrix is $mp$ x $p$.  Since $n^{-1} \mathbf{X}_{\mathcal{S}_1}^T \mathbf{X}_{\mathcal{S}_1}$ is an empirical covariance matrix (maximum likelihood estimate of $\mathbf{R}_{\mathcal{S}_1,\mathcal{S}_1}$), we denote the true inverse covariance matrix of signals from the active set by $\mathbf{G}_{\mathcal{S}_1} = \mathbf{R}_{\mathcal{S}_1,\mathcal{S}_1}^{-1} = [\begin{smallmatrix} \mathbf{G}_{\mathcal{S}_{1,1}} & \ldots & \mathbf{G}_{\mathcal{S}_{1,m}} \end{smallmatrix}]$.

%Let $\sigma_{\mathcal{A}_i}$ denote the maximum singular value of matrix $\mathbf{G}_{\mathcal{A}_i}$ and $\sigma_\mathcal{A}$ denote the maximum singular value (i.e. eigenvalue) of $\mathbf{G}_\mathcal{A}$.  Note $C_{min}^{-1} \geq \sigma_\mathcal{A} \geq \max_i \sigma_{\mathcal{A}_i}$.

To show that each subvector $\hat{\mathbf{a}}_{1,i}^{*} \neq 0$ for $i \in \mathcal{S}_1$ in the limit, it suffices to show that $\| \widehat{\mathbf{G}}_{\mathcal{S}_{1,i}}^T (\frac{\lambda}{2} \hat{\mathbf{z}}_{\mathcal{S}_1}  - \frac{1}{n} \mathbf{X}_{\mathcal{S}_1}^T \mathbf{u}_1 ) \|_2 < C_{con} \leq \| \mathbf{a}_{\mathcal{S}_{1,i}} \|_2$.  Applying the triangle inequality, we instead show that $\| \widehat{\mathbf{G}}_{\mathcal{S}_1}^T (\frac{\lambda}{2} \hat{\mathbf{z}}_{\mathcal{S}_1}  - \frac{1}{n} \mathbf{X}_{\mathcal{S}_1}^T \mathbf{u}_1 ) \|_2 < C_{con}$ with the following lemma.

%than each subvector will also have norm less than $C_{con}$.

\begin{lemma}
\label{lma:oracleres}
Given Assumptions~\ref{as:scaling}--\ref{as:maxsv}, $\| \widehat{\mathbf{G}}_{\mathcal{S}_1}^T (\frac{\lambda}{2} \hat{\mathbf{z}}_{\mathcal{S}_1}  - \frac{1}{n} \mathbf{X}_{\mathcal{S}_1}^T \mathbf{u}_1 ) \|_2 = \mathcal{O}(\max{(n^{\frac{c_2-c_4}{2}}, n^{c_2+\frac{c_3-c_4-1}{2}}, n^{-\frac{1}{2}})})$ with probability exceeding $1-\exp{(-\Theta(n))}$.
\end{lemma}

\begin{IEEEproof}
Using $\| \widehat{\mathbf{G}}_{\mathcal{S}_{1}}^T (\frac{\lambda}{2} \hat{\mathbf{z}}_{\mathcal{S}_1}  - \frac{1}{n} \mathbf{X}_{\mathcal{S}_1}^T \mathbf{u}_1 ) \|_2 \leq \frac{\lambda}{2} \| \widehat{\mathbf{G}}_{\mathcal{S}_{1}}^T \hat{\mathbf{z}}_{\mathcal{S}_1} \|_2 + \frac{1}{n} \| \widehat{\mathbf{G}}_{\mathcal{S}_{1}}^T \mathbf{X}_{\mathcal{S}_1}^T \mathbf{u}_1 \|_2$, we bound the two terms separately.  First:

\begin{eqnarray*}
\lefteqn{ \frac{\lambda}{2} \| \widehat{\mathbf{G}}_{\mathcal{S}_1}^T \hat{\mathbf{z}}_{\mathcal{S}_1} \|_2 \leq \frac{\lambda}{2} \| \widehat{\mathbf{G}}_{\mathcal{S}_1} \|_2 \| \hat{\mathbf{z}}_{\mathcal{S}_1} \|_2 } \\
& \leq & \frac{\lambda \sqrt{m}}{2} \| \widehat{\mathbf{G}}_{\mathcal{S}_1} \|_2 \\
& \leq & \frac{\lambda \sqrt{m}}{2} \left( \| \mathbf{G}_{\mathcal{S}_1} \|_2 + \| \widehat{\mathbf{G}}_{\mathcal{S}_1} - \mathbf{G}_{\mathcal{S}_1} \|_2 \right) \\
& \leq & \frac{\lambda \sqrt{m}}{2} \left( C_{min}^{-1} + \| \mathbf{G}_{\mathcal{S}_1} \|_2 \left\| \left( \left(\frac{\mathbf{W}^T\mathbf{W}}{n}\right)^{-1} - \mathbf{I} \right) \right\|_2 \right) \\
& < & \frac{\lambda \sqrt{m}}{2} \left( C_{min}^{-1} + \mathcal{O} \left( \sqrt{\frac{mp}{n}} \right) \right) \\
& < & \mathcal{O}(n^{(c_2-c_4)/2})+\mathcal{O}(n^{c_2-c4/2+c_3/2-1/2})
\end{eqnarray*}

\noindent
where $\mathbf{W} \sim \mathcal{N}(\mathbf{0},\mathbf{I})$.  The second inequality is simply the triangle inequality applied to $\hat{\mathbf{z}}_{\mathcal{S}_1}$ since $\| \hat{\mathbf{z}}_{1,i} \|_2 = 1$ for all $i \in \mathcal{S}_1$ and each node has no more than $m$ parents.  The second to last inequality holds with probability greater than $1-\exp{(-\Theta(n))}$ \cite{obozinski:08}.  Given the conditions on constants $c_2$--$c_4$, the last line goes to zero.

Next, consider:

\begin{eqnarray}
\frac{1}{n} \| \widehat{\mathbf{G}}_{\mathcal{S}_1}^T \mathbf{X}_{\mathcal{S}_1}^T \mathbf{u}_1 \|_2 & = & \| (\mathbf{X}_{\mathcal{S}_1}^T \mathbf{X}_{\mathcal{S}_1})^{-1} \mathbf{X}_{\mathcal{S}_1}^T \mathbf{u}_1 \|_2 \nonumber \\
& = & \| (\mathbf{X}_{\mathcal{S}_1}^+)^{T} \mathbf{u}_1 \|_2 \nonumber \\
& \leq & \frac{\sigma_1^2}{n} \left\| \widehat{\mathbf{G}}_{\mathcal{S}_1} \right\|_2^{1/2} \| \mathbf{u}_1 \|_2
\label{eq:chibound1}
\end{eqnarray}

%\begin{equation}
% \label{eq:chibound1}
%\| \widehat{\mathbf{G}}_{\mathcal{S}_1}^T \mathbf{X}_{\mathcal{S}_1}^T \mathbf{u}_1 \|_2 = \sigma_{1}^2 \left\| \widehat{\mathbf{G}}_{\mathcal{S}_1}^{1/2} \tilde{\mathbf{u}}_1 \right\|_2 \leq \sigma_{1}^2 \left\| \widehat{\mathbf{G}}_{\mathcal{S}_1} \right\|_2^{1/2} \| \tilde{\mathbf{u}}_1 \|_2
%\end{equation}

\noindent
where $\mathbf{X}_{\mathcal{S}_1}^+$ denotes the pseudoinverse and $\mathbf{u}_1 \sim \mathcal{N}(\mathbf{0},\sigma_1^2 \mathbf{I})$ since we have assumed independent time samples.  Inequality~(\ref{eq:chibound1}) can be easily seen by considering the singular value decomposition of $\mathbf{X}_{\mathcal{S}_1}$.  Obozinski et al. \cite{obozinski:08} provide the following bound for the inverse sample covariance matrix:

\begin{equation*}
\Pb\left( \| \widehat{\mathbf{G}}_{\mathcal{S}_1} \|_2 \leq 2 C_{min}^{-1}\right) \geq 1 - 2 \exp(-\Theta(n))
\end{equation*}

\noindent
and \cite{laurent:00} provide a bound for the chi-square variate:

\begin{equation*}
\Pb\left(\| \tilde{\mathbf{u}}_1 \|_2^2 - n \geq 2 \sqrt{nt} + 2t \right) \leq \exp{(-t)}
\end{equation*}

\noindent
which holds for any $t>0$.  In particular, $\| \tilde{\mathbf{u}}_1 \|_2^2 < 5n$ for $t=n$ with probability exceeding $1-\exp{(-n)}$.  Combining these bounds with (\ref{eq:chibound1}) and Assumption~\ref{as:dnoise} gives us:

\begin{equation*}
\frac{1}{n} \| \widehat{\mathbf{G}}_{\mathcal{S}_{1,i}}^T \mathbf{X}_{\mathcal{S}_1}^T \mathbf{u}_1 \|_2 < \frac{C_{power}}{\sqrt{n}} \sqrt{10 C_{min}^{-1}} = \mathcal{O}\left( \frac{1}{\sqrt{n}} \right)
\end{equation*}

\noindent
with probability greater than $1 - 2 \exp{(-\Theta(n))}$.

\end{IEEEproof}

Since both terms of $\| \widehat{\mathbf{G}}_{\mathcal{S}_1}^T (\frac{\lambda}{2} \hat{\mathbf{z}}_1  - \frac{1}{n} \mathbf{X}_{\mathcal{S}_1}^T \mathbf{u}_1 ) \|_2$ go to zero as $n$ grows, their sum will be less than $C_{con}$ with high probability for large $n$.  This implies that each $\| \widehat{\mathbf{G}}_{\mathcal{S}_{1,i}}^T (\frac{\lambda}{2} \hat{\mathbf{z}}_1  - \frac{1}{n} \mathbf{X}_{\mathcal{S}_1}^T \mathbf{u}_1 ) \|_2$, $i \in \mathcal{S}_1$ will also be less than $C_{con}$, so for all $i \in \mathcal{S}_1$, $\| \hat{\mathbf{a}}_{1,i}^{*} \|_2 > \| \mathbf{a}_{1,i} \|_2 - C_{con} \geq 0$.

We have shown that $\hat{\mathbf{a}}_{1,i}^{*} \neq 0$ for each $i \in \mathcal{S}_1$ with probability greater than $1 - \exp{(-\Theta(n))}$.  We next show that the oracle solution is in fact the overall solution with high probability.

\subsection*{Limiting False Positives}
%\label{limfp}

Assuming that the oracle solution from (\ref{eq:oracleincl}) has all nonzero subvectors $\hat{\mathbf{a}}_{1,i}^{*}$, we must ensure that $\hat{\mathbf{a}}^* = [\begin{smallmatrix} (\hat{\mathbf{a}}_{\mathcal{S}_1}^{*})^T & \mathbf{0}^T \end{smallmatrix}]^T$ is a solution to the full problem with high probability.  In other words, we must show that $\frac{2}{\lambda n} \| \mathbf{X}_j^T (\mathbf{y} - \mathbf{X} \hat{\mathbf{a}}^*) \|_2 \leq 1$ for all $j \in \mathcal{S}_1^C$.  To do so, we adopt a technique used in \cite{meins:06}.  Write $\mathbf{X}_j = \sum_{i \in \mathcal{S}_1} \mathbf{X}_i \Psib_{j,i} + \mathbf{V}_j$, where

\begin{equation}
\Psib_{j,\mathcal{S}_1} = \begin{bmatrix} \Psib_{j,1} \\ \vdots \\ \Psib_{j,m} \end{bmatrix} = \arg \min \E\left[ \left\| \mathbf{X}_j - \sum_{i \in \mathcal{S}_1} \mathbf{X}_i \Psib_{j,i} \right\|_F^2 \right],
\label{eq:bigpsi}
\end{equation}

\noindent
and $\mathbf{V}_j$ is a random variable representing the portion of $\mathbf{X}_j$ that can't be predicted by $\mathbf{X}_i$, $i \in \mathcal{S}_1$.  Now we have:

\begin{eqnarray}
\lefteqn{ \frac{2}{\lambda n} \left\| \mathbf{X}_j^T (\mathbf{y}_1 - \mathbf{X} \hat{\mathbf{a}}_1^*) \right\|_2 } \\
& = & \frac{2}{\lambda n} \left\| \left(\sum_{i \in \mathcal{S}_1} \mathbf{X}_i \Psib_{j,i} + \mathbf{V}_j\right)^T (\mathbf{y}_1 - \mathbf{X} \hat{\mathbf{a}}_1^*) \right\|_2 \nonumber \\
& = & \frac{2}{\lambda n} \left\| \sum_{i \in \mathcal{S}_1} \Psib_{j,i}^T  \mathbf{X}_i^T (\mathbf{y}_1 - \mathbf{X} \hat{\mathbf{a}}_1^*) + \mathbf{V}_j^T (\mathbf{y}_1 - \mathbf{X} \hat{\mathbf{a}}_1^*) \right\|_2 \nonumber \nonumber \\
& = & \left\| \sum_{i \in \mathcal{S}_i} \Psib_{j,i}^T  \frac{\hat{\mathbf{a}}_{1,i}^*}{\| \hat{\mathbf{a}}_{1,i}^* \|_2}+ \frac{2}{\lambda n} \mathbf{V}_j^T (\mathbf{y}_1 - \mathbf{X} \hat{\mathbf{a}}_1^*) \right\|_2 \label{eq:excldecomp} \\
& \leq & \left\| \sum_{i \in \mathcal{S}_i} \Psib_{j,i}^T \left( \frac{\hat{\mathbf{a}}_{1,i}^*}{\| \hat{\mathbf{a}}_{1,i}^* \|_2} - \frac{\mathbf{a}_{1,i}^*}{\| \mathbf{a}_{1,i}^* \|_2} \right) \right\|_2 \nonumber \\
& + & \left\| \sum_{i \in \mathcal{S}_1} \Psib_{j,i}^T \frac{\mathbf{a}_{1,i}^*}{\| \mathbf{a}_{1,i}^* \|_2} \right\|_2 + \frac{2}{\lambda n} \left\| \mathbf{V}_j^T (\mathbf{y}_1 - \mathbf{X} \hat{\mathbf{a}}_1^*) \right\|_2 \label{eq:limfpcon}
\end{eqnarray}

\noindent
where (\ref{eq:excldecomp}) follows from the KKT condition (\ref{eq:inclcond}).  The second term of (\ref{eq:limfpcon}) is less than one by Assumption~\ref{as:represent}.  We bound the remaining terms separately. In order to bound the first term, we use the following lemma:

%$\| \mathbf{v} - \mathbf{w} \| < \frac{\epsilon}{2}$ implies

\begin{lemma}
\label{lma:dist2angle}
$\left\| \frac{\mathbf{v}}{\| \mathbf{v} \|} - \frac{\mathbf{w}}{\| \mathbf{w} \|} \right\| < \frac{2 \| \mathbf{v} - \mathbf{w} \| }{\| \mathbf{w} \|}$
\end{lemma}

\begin{IEEEproof}
%We show $\left\| \frac{\mathbf{v}}{\| \mathbf{v} \|} - \frac{\mathbf{w}}{\| \mathbf{w} \|} \right\| < \frac{\epsilon}{\| \mathbf{w} \|}$ by bounding two terms:

\begin{eqnarray*}
\left\| \frac{\mathbf{v}}{\| \mathbf{v} \|} - \frac{\mathbf{w}}{\| \mathbf{w} \|} \right\| & \leq & \left\| \frac{\mathbf{v}}{\| \mathbf{v} \|} - \frac{\mathbf{v}}{\| \mathbf{w} \|} \right\| + \left\| \frac{\mathbf{v}}{\| \mathbf{w} \|} - \frac{\mathbf{w}}{\| \mathbf{w} \|} \right\| \\
& = & \| \mathbf{v} \| \left| \frac{1}{\| \mathbf{v} \|} - \frac{1}{\| \mathbf{w} \|} \right| + \frac{\| \mathbf{v} - \mathbf{w} \|}{\|\mathbf{w}\|}  \\
& \leq & \frac{\left| \left(\| \mathbf{w} \| - \| \mathbf{v} \| \right) \right|}{\| \mathbf{w} \|}   + \frac{\| \mathbf{v} - \mathbf{w} \|}{\|\mathbf{w}\|} \\
& \leq & \frac{2 \| \mathbf{v} - \mathbf{w} \|}{\|\mathbf{w}\|} \\
\end{eqnarray*}
\end{IEEEproof}

% & \leq & \frac{\epsilon}{\| \mathbf{w} \|}

We now bound the first term of (\ref{eq:limfpcon}):

\begin{eqnarray*}
\lefteqn{ \left\| \sum_{i \in \mathcal{S}_1} \Psib_{j,i}^T \left( \frac{\hat{\mathbf{a}}_{1,i}^*}{\| \hat{\mathbf{a}}_{1,i}^* \|_2} - \frac{\mathbf{a}_{1,i}}{\| \mathbf{a}_{1,i} \|_2} \right) \right\|_2} \nonumber \\
& \leq & \| \Psib_{j,\mathcal{S}_1} \|_2 \left( \sum_{i \in \mathcal{S}_1} \left\| \frac{\hat{\mathbf{a}}_{1,i}^*}{\| \hat{\mathbf{a}}_{1,i}^* \|_2} - \frac{\mathbf{a}_{1,i}}{\| \mathbf{a}_{1,i} \|_2} \right\|_2^2 \right)^{1/2} \nonumber \\
& \leq & \| \Psib_{j,\mathcal{S}_1} \|_2 \left( \sum_{i \in \mathcal{S}_1} \frac{2 \left\| \hat{\mathbf{a}}_{1,i}^* - \mathbf{a}_{1,i} \right\|_2^2}{\| \mathbf{a}_{1,i} \|_2^2} \right)^{1/2} \nonumber \\
\end{eqnarray*}

\noindent
where we have applied Lemma~\ref{lma:dist2angle}.  From Assumption~\ref{as:cstrength} we have $\| \mathbf{a}_{1,i} \|_2 \geq C_{con}$.  Using this and $\| \Psib_{j,\mathcal{S}_1} \|_2 = \| \mathbf{R}_{\A_1,\A_1}^{-1} \E[\mathbf{X}_{\A_1}^T \mathbf{X}_j] \|_2 \leq  \| \mathbf{R}_{\A_1,\A_1}^{-1} \mathbf{R}_{\A_1,\A_1^C} \|_2 \leq C_{max}C_{min}^{-1}$, we have:

\begin{eqnarray*}
\lefteqn{ \left\| \sum_{i \in \mathcal{S}_1} \Psib_{j,i}^T \left( \frac{\hat{\mathbf{a}}_{1,i}^*}{\| \hat{\mathbf{a}}_{1,i}^* \|_2} - \frac{\mathbf{a}_{1,i}}{\| \mathbf{a}_{1,i} \|_2} \right) \right\|_2} \nonumber \\
& \leq & \sqrt{2} \| \Psib_{j} \|_2 C_{con}^{-1} \left( \sum_{i \in \mathcal{S}_1}  \left\| \hat{\mathbf{a}}_{1,i}^* - \mathbf{a}_{1,i} \right\|_2^2 \right)^{1/2} \nonumber \\
& \leq & \sqrt{2} \frac{C_{max}}{C_{min} C_{con}} \left\| \hat{\mathbf{a}}_{\mathcal{S}_1}^{*} - \mathbf{a}_{\mathcal{S}_1} \right\|_2 \nonumber \\
& = & \mathcal{O}(\max{(n^{\frac{c_2-c_4}{2}}, n^{c_2+\frac{c_3-c_4-1}{2}}, n^{-\frac{1}{2}})})
\end{eqnarray*}

\noindent
where the last inequality follows from (\ref{eq:ahatoracle}) and Lemma~\ref{lma:oracleres}.

%Using Lemma~\ref{lma:dist2angle} and Assumption~\ref{as:cstrength} gives us $\left\| \frac{\hat{\mathbf{a}}_{\mathcal{S}}^{*}}{\| \hat{\mathbf{a}}_{\mathcal{S}}^{*} \|_2} - \frac{\mathbf{a}_{\mathcal{S}}}{\| \mathbf{a}_{ \mathcal{S}} \|_2} \right\|_2 \leq \frac{2 \epsilon(n)}{\| \mathbf{a}_{ \mathcal{S}} \|_2} \leq \frac{2 \epsilon(n)}{C_{con} \sqrt{m}}$.  Turning back to the first term of (\ref{eq:limfpcon}), we have, for vector $\mathbf{v} = [\begin{smallmatrix} \mathbf{v}_1^T \ldots \mathbf{v}_m^T \end{smallmatrix}]^T$ with norm no greater than $2 \epsilon(n) C_{con}^{-1} m^{-1/2}$:

Finally, we show that the last term of (\ref{eq:limfpcon}) goes to zero faster than $\mathcal{O}(n^{(c_3+c_4-1)/2})$.  Since they are linear combinations of zero mean Gaussian random vectors, the $p$ columns of $\mathbf{V}_j$ as well as vector $\mathbf{y}-\mathbf{X} \hat{\mathbf{a}}^{*}$ are Gaussian.  Though these $p+1$ vectors will be correlated for most interesting networks, the entries in any one of these vectors are i.i.d. Gaussian with variance less than $C_{power}$.  We establish the following lemma.

\begin{lemma}
\label{lma:chibound}
Let $\mathbf{V}$ be an $n$ by $p$ random matrix and $\mathbf{w}$ an $n$ dimensional random vector.  For each $i = 1,2,\ldots,n$, let the $i^{th}$ row of $\mathbf{V}$ concatenated with the $i^{th}$ entry of $\mathbf{w}$ be i.i.d. Gaussian vectors with distribution $\mathcal{N}(\mathbf{0},\mathbf{C})$, for some covariance matrix $\mathbf{C}$ whose maximum (diagonal) entry is $C_m$.  Then with probability exceeding $1-p \exp(-n)$, $\| \mathbf{V}^T \mathbf{w} \|_2 < C_m \sqrt{5np}$.
\end{lemma}

\begin{IEEEproof}
The entries in any column of $\mathbf{V}$ are i.i.d. Gaussian with variance less than $C_m$, as are the entries of $\mathbf{w}$.  With this in mind, we bound each entry of $\mathbf{z} \equiv \mathbf{V}^T \mathbf{w}$ by $C_m$ times a chi-squared random variable with $n$ degrees of freedom (denoted $\tilde{z}_i \sim \chi_n^2$ for $i=1,2,\ldots,p$) and use the union bound:

% changed a (typo) $\mathbf{v}$ to $\mathbf{w}$ above, 3/31/2010

\begin{eqnarray*}
\Pb(\| \mathbf{z} \|_2^2 \geq C_m^2 5np) & \leq & \Pb\left(\| \tilde{\mathbf{z}} \|_2^2 \geq 5np \right) \\
& \leq & p \Pb\left( \tilde{z}_1^2 \geq 5n\right) \\
& \leq & p \exp(-n)
\end{eqnarray*}

\noindent
where we have used the same chi-squared bound as in Lemma~\ref{lma:oracleres}.  Thus with probability exceeding $1-p \exp(-n)$, $\| \mathbf{V}^T \mathbf{w} \|_2 < C_m \sqrt{5np}$.

\end{IEEEproof}

Using Lemma~\ref{lma:chibound}, we have with probability exceeding $1-p \exp(-n)$, $\| \mathbf{V}_j^T (\mathbf{y} - \mathbf{X} \hat{\mathbf{a}}^{*}) \|_2 < C_{power} \sqrt{5np}$.  Dividing by $\lambda n$ and using Assumption~\ref{as:scaling}, we have:

\begin{equation}
\frac{2}{\lambda n} \| \mathbf{V}_j^T (\mathbf{y} - \mathbf{X} \hat{\mathbf{a}}^{*}) \|_2 < \frac{C_{power} \sqrt{5p}}{\lambda \sqrt{n}} = \mathcal{O}(n^{(c_3+c_4-1)/2}).
\label{eq:resbound}
\end{equation}

By (\ref{eq:exclcond}), there will be no false positives if (\ref{eq:limfpcon}) is less than one.  With high probability, the second term is less than $C_{fc} < 1$ by Assumption~\ref{as:represent}.  The first and third terms go to zero with large $n$ with high probability.

\subsection*{Union Bound}

We have shown that (\ref{eq:glasso_ar}) recovers the correct parents of node~$1$ (set $\mathcal{S}_1$) with probability exceeding $1-\exp{(-\Theta(n))}$.  To obtain the result for the whole network, we apply the union bound:

\begin{eqnarray*}
\Pb\left( \bigcup_{i=1}^N \hat{\mathcal{S}}_i \neq \mathcal{S}_i\right) & \leq & \sum_{i=1}^N \Pb\left( \hat{\mathcal{S}}_i \neq \mathcal{S}_i\right) \\
& \leq & N \exp{(-\Theta(n))} \\
& \leq & n^{c_1} \exp{(-\Theta(n))} \\
& \leq & \exp{(c_1 \ln{n} - \Theta(n))} \\
& \leq & \exp{(-\Theta(n))} \\
\end{eqnarray*}

\section{Proof of Necessary Condition}
\label{GSMAR:neccond}

We must show that (\ref{eq:glasso_ar}) will not recover the correct set of nonzero $\mathbf{a}_{1,i}$ when Assumptions~\ref{as:dnoise}--\ref{as:maxsv} hold but $\left\| \sum_{i \in \mathcal{S}_1} \Psib_{j,i}^T  \frac{\mathbf{a}_{1,i}}{\| \mathbf{a}_{1,i} \|_2} \right\|_2 > 1 + c$.  We do so by contradiction.

Suppose $\lambda$ scales with $n$ such that all the coefficient blocks in $\A_1$ of the oracle solution are nonzero and the probability of false positives goes to zero as $n$ grows.  Then KKT condition (\ref{eq:exclcond}) must hold with high probability for large $n$.  This implies the following bound must hold with high probability for all $j \in \A_1^C$:

\begin{eqnarray}
\frac{\lambda}{2} & \geq & n^{-1} \| \mathbf{X}_{j}^T (\mathbf{y}_1 - \mathbf{X} \hat{\mathbf{a}}_1^*) \|_2 \nonumber \\
& = & \frac{\lambda}{2} \left\| \sum_{i \in \mathcal{S}_1} \Psib_{j,i}^T  \frac{\hat{\mathbf{a}}_{1,i}^{*}}{\| \hat{\mathbf{a}}_{1,i}^{*} \|_2} + \frac{2}{\lambda n} \mathbf{V}_j^T (\mathbf{y}_1 - \mathbf{X} \mathbf{a}_1^{*}) \right\|_2 \nonumber \\
& \geq & \frac{\lambda}{2} \left\| \sum_{i \in \mathcal{S}_1} \Psib_{j,i}^T  \frac{\mathbf{a}_{1,i}}{\| \mathbf{a}_{1,i} \|_2} \right\|_2 - \frac{\lambda}{2}  \| \Psib_j^T \mathbf{w} \|_2 \nonumber \\
& & - n^{-1} \left\| \mathbf{V}_j^T (\mathbf{y}_1 - \mathbf{X} \mathbf{a}_{1}^{*}) \right\|_2 \nonumber \\
& > & \frac{\lambda}{2}(1+c) - \frac{\lambda}{2} \left\| \Psib_j^T \mathbf{w} \right\|_2 - n^{-1} \left\| \mathbf{V}_j^T (\mathbf{y}_1 - \mathbf{X} \mathbf{a}_1^{*}) \right\|_2 \nonumber \\
\label{eq:lb}
\end{eqnarray}

\noindent
where $\mathbf{w}=[\begin{smallmatrix} \mathbf{w}_1 \ldots \mathbf{w}_m \end{smallmatrix}]^T$ and $\mathbf{w}_{1,i} = \left(\frac{\hat{\mathbf{a}}_{1,i}^*}{\| \hat{\mathbf{a}}_{1,i}^* \|} -  \frac{\mathbf{a}_{1,i}}{\| \mathbf{a}_{1,i} \|}\right)$.  From Eq.~(\ref{eq:resbound}) we have $n^{-1} \left\| \mathbf{V}_j^T (\mathbf{y}_1 - \mathbf{X} \mathbf{a}_1^{*}) \right\|_2 = \mathcal{O}(\sqrt{p/n})$, which goes to zero since $p/n \leq mp/n$, which goes to zero as $n$ goes to infinity by assumption.  We have also shown that $\left\| \Psib_j^T \mathbf{w} \right\|_2$ goes to zero; however, this term is now multiplied by $\frac{\lambda}{2}$ for some unknown $\lambda$ scaling.  To proceed, Eq.~(\ref{eq:lb}) implies:

\begin{equation*}
\frac{c \lambda}{2} < \frac{\lambda}{2} \left\| \Psib_j^T \mathbf{w} \right\|_2 + \mathcal{O}(\sqrt{p/n})
\end{equation*}

\noindent
Since the second term goes to zero, this implies:

% START NOTATION FIX HERE

\begin{equation*}
c < \| \Psib_j^T \mathbf{w} \|_2 \leq \left\| \Psib_{j} \right\|_2 \left\| \mathbf{w} \right\|_2 \leq \frac{C_{max}}{C_{min}} \sqrt{m} \max_i \left\| \mathbf{w}_i \right\|_2
\end{equation*}

\noindent
where the last inequality follows from the definition of $\Psib_j$ and the triangle inequality.  This means there is at least one $i \in \mathcal{S}_1$ for which $\left\| \frac{\hat{\mathbf{a}}_{1,i}^*}{\| \hat{\mathbf{a}}_{1,i}^* \|_2} - \frac{\mathbf{a}_{1,i}}{\| \mathbf{a}_{1,i} \|_2} \right\|_2 \geq \frac{c C_{min}}{\sqrt{m} C_{max}}$.  Combining this with Lemma~(\ref{lma:dist2angle}) implies that $\left\| \hat{\mathbf{a}}_{1,i} - \mathbf{a}_{1,i} \right\|_2 \geq \frac{c C_{min} \| \mathbf{a}_{1,i} \|_2}{2\sqrt{m} C_{max}}$.

Now we use Assumption~\ref{as:cstrength} and (\ref{eq:ahatoracle}):

\begin{eqnarray*}
\frac{c C_{min} C_{con}}{2 \sqrt{m} C_{max}} & \leq & \frac{c C_{min} \| \mathbf{a}_{1,i} \|_2}{2 \sqrt{m} C_{max}} \\
& \leq & \| \mathbf{a}_{1,i} - \hat{\mathbf{a}}_{1,i} \|_2 \\
& = & \left\| \widehat{\mathbf{G}}_{\mathcal{S}_{1,i}}^T \left(\frac{\lambda}{2} \hat{\mathbf{z}}_{\mathcal{S}_1}  - \frac{1}{n} \mathbf{X}_{\mathcal{S}_1}^T \mathbf{u}_1 \right) \right\|_2 \\
& = & \left\| \begin{bmatrix} \mathbf{I}_p & \mathbf{0} \end{bmatrix} \widehat{\mathbf{G}}_{\mathcal{S}_{1}}^T \left(\frac{\lambda}{2} \hat{\mathbf{z}}_{\mathcal{S}_1}  - \frac{1}{n} \mathbf{X}_{\mathcal{S}_1}^T \mathbf{u}_1 \right) \right\|_2 \\
& \leq & \left\| \widehat{\mathbf{G}}_{\mathcal{S}_1}^T \left(\frac{\lambda}{2} \hat{\mathbf{z}}_{\mathcal{S}_1} - \frac{1}{n} \mathbf{X}_{\mathcal{S}_1}^T \mathbf{u}_1 \right) \right\|_2 \\
& < & \frac{\lambda \sqrt{m}}{2} \left( C_{min}^{-1} + \mathcal{O}\left( \sqrt{\frac{m p}{n}} \right) \right)
\end{eqnarray*}

\noindent
where the last inequality follows from the proof of Lemma~\ref{lma:oracleres}.  Since $m p / n$ goes to zero, we have the following lower bound on $\lambda$:

\begin{equation}
\label{eq:lambdalb}
\lambda > \frac{c C_{min}^2 C_{con}}{m C_{max}}
\end{equation}

Since $\hat{\mathbf{a}}_{1,i} \neq 0$ for at least one $i$ by assumption, KKT condition (\ref{eq:inclcond}), repeated here for readability, must hold for at least one $i$:

\begin{eqnarray}
\mathbf{X}_i^T (\mathbf{y}_1 - \mathbf{X} \hat{\mathbf{a}}_1) = \frac{\lambda n \hat{\mathbf{a}}_{1,i}}{2 \| \hat{\mathbf{a}}_{1,i} \|_2} & & \forall \ i \: \mathrm{s.t.} \: \hat{\mathbf{a}}_{1,i} \neq \mathbf{0}
\label{eq:repeatincl}
\end{eqnarray}

\noindent
Using Lemma~\ref{lma:chibound} (with $\mathbf{V} = \mathbf{X}_i$ and $\mathbf{w} = \mathbf{y}_1 - \mathbf{X} \hat{\mathbf{a}}_1$), the norm of the left hand side of (\ref{eq:repeatincl}) is less than $C_{power} \sqrt{5 n p}$ with high probability for large $n$.  On the other hand, (\ref{eq:lambdalb}) implies that the norm of the right hand side of (\ref{eq:repeatincl}) is $\Omega(n/m)$.  Given that $n$ grows faster than $m^2 p$, this is a contradiction.

The scaling law $n>m^2 p$ for large $n$ (equivalently $2c_2 + c_3 <1$) was not required to prove asymptotic consistency.  Other proof techniques may result in matching scaling laws.

%Assuming (\ref{eq:glasso_ar})

\section{Proof of Corollary \ref{cor:freeparam}}
\label{app:corollaryproof}

The proof is the same as that of Theorem~\ref{thm:ascon} with a few minor changes.  The KKT condition (\ref{eq:inclcond}) for $l=1$ becomes $\mathbf{X}_1^T (\mathbf{y}_1 - \mathbf{X} \hat{\mathbf{a}}_1) = \mathbf{0}$, which implies $\hat{\mathbf{z}}_{1,1}=\mathbf{0}$.  The results of Lemma~\ref{lma:oracleres} still apply with $m$ replaced by $m-1$ in the proof.  In App.~\ref{GSMAR:asymptcons}, $\psi_{j \rightarrow 1}^{FC} = \left\| \sum_{i \in \mathcal{S}_1} \Psib_{j,i}^T  \frac{\hat{\mathbf{a}}_{1,i}}{\| \hat{\mathbf{a}}_{1,i} \|_2} \right\|_2$ is simply replaced with $\tilde{\psi}_{j \rightarrow 1}^{FC} = \left\| \sum_{i \in \mathcal{S}_1,i \neq 1} \Psib_{j,i}^T  \frac{\hat{\mathbf{a}}_{1,i}}{\| \hat{\mathbf{a}}_{1,i} \|_2} \right\|_2$ since $\mathbf{X}_1^T (\mathbf{y}_1 - \mathbf{X} \hat{\mathbf{a}}_1) = \mathbf{0}$ instead of $\frac{\lambda n \hat{\mathbf{a}}_{1,1}}{2 \| \hat{\mathbf{a}}_{1,1} \|_2}$.

%\section*{Acknowledgment}

%Thank you to Wim van Drongelen and Hyong Lee at University of Chicago Hospitals for sharing the ECoG data and sharing their expertise in its analysis.

\ifCLASSOPTIONcaptionsoff
  \newpage
\fi



\bibliographystyle{IEEEtran}
%\bibliography{IEEEabrv,../../PhDpaper}
%\bibliography{IEEEabrv,netsbib}
\bibliography{netsbib}

\begin{thebibliography}{10}
\providecommand{\url}[1]{#1}
\csname url@samestyle\endcsname
\providecommand{\newblock}{\relax}
\providecommand{\bibinfo}[2]{#2}
\providecommand{\BIBentrySTDinterwordspacing}{\spaceskip=0pt\relax}
\providecommand{\BIBentryALTinterwordstretchfactor}{4}
\providecommand{\BIBentryALTinterwordspacing}{\spaceskip=\fontdimen2\font plus
\BIBentryALTinterwordstretchfactor\fontdimen3\font minus
  \fontdimen4\font\relax}
\providecommand{\BIBforeignlanguage}[2]{{%
\expandafter\ifx\csname l@#1\endcsname\relax
\typeout{** WARNING: IEEEtran.bst: No hyphenation pattern has been}%
\typeout{** loaded for the language `#1'. Using the pattern for}%
\typeout{** the default language instead.}%
\else
\language=\csname l@#1\endcsname
\fi
#2}}
\providecommand{\BIBdecl}{\relax}
\BIBdecl

\bibitem{baccala:01}
L.~Baccal{\'{a}} and K.~Sameshima, ``Partial directed coherence: a new concept
  in neural structure determination,'' \emph{Biological Cybernetics}, vol.~84,
  pp. 463--474, 2001.

\bibitem{kaminski:05}
M.~Kami{\'{n}}ski, ``Determination of transmission patterns in multichannel
  data,'' \emph{Philosophical Transactions of the Royal Society B}, vol. 360,
  pp. 947--952, 2005.

\bibitem{winterhalder:05}
M.~Winterhalder, B.~Schelter, W.~Hesse, K.~Schwab, L.~Leistritz, D.~Klan,
  R.~Bauer, J.~Timmer, and H.~Witte, ``Comparisson of linear signal processing
  techniques to infer directed interactions in multivariate neural systems,''
  \emph{Signal Processing}, vol.~85, no.~11, pp. 2137--2160, 2005.

\bibitem{lutkepohl:91}
H.~L{\"{u}}tkepohl, \emph{Introduction to Multiple Time Series Analysis}.\hskip
  1em plus 0.5em minus 0.4em\relax Berlin: Springer-Verlag, 1991.

\bibitem{tibshirani:96}
R.~Tibshirani, ``Regression shrinkage and selection via the lasso,''
  \emph{Journal of the Royal Statistical Society Series B}, vol.~58, pp.
  267--288, 1996.

\bibitem{yuan:06}
M.~Yuan and Y.~Lin, ``Model selection and estimation in regression with grouped
  variables,'' \emph{Journal of the Royal Statistical Society Series B},
  vol.~68, no.~1, pp. 49--67, 2006.

\bibitem{akb:07}
A.~Bolstad, B.~Van~Veen, and R.~Nowak, ``Space-time sparsity regularization for
  the magnetoencephalography inverse problem,'' in \emph{4th IEEE International
  Symposium on Biomedical Imaging}, Arlington, VA, April 2007, pp. 984--987.

\bibitem{akb:ssp07}
------, ``Magneto-/electroencephalography with space-time sparse priors,'' in
  \emph{IEEE Statistical Signal Processing Workshop}, Madison, WI, August 2007,
  pp. 190--194.

\bibitem{ding_he:07}
L.~Ding and B.~He, ``Sparse source imaging in {EEG},'' in \emph{Proceedings of
  NFSI \& ICFBI}, Hangzhou, China, October 2007.

\bibitem{ou:08}
W.~Ou, M.~H{\"{a}}m{\"{a}}l{\"{a}}inen, and P.~Golland, ``A distributed
  spatio-temporal eeg/meg inverse solver,'' \emph{NeuroImage}, vol.~44, no.~3,
  pp. 932--946, 2009.

\bibitem{haufe:08}
S.~Haufe, V.~Nikulin, A.~Ziehe, K.-R. M{\"{u}}ller, and G.~Nolte, ``Combining
  sparsity and rotational invariance in eeg/meg source reconstruction,''
  \emph{NeuroImage}, vol.~42, no.~2, pp. 726--738, 2008.

\bibitem{akb_ni:09}
A.~Bolstad, B.~van Veen, and R.~Nowak, ``Space-time event sparse penalization
  for mangeto-/electroencephalography,'' \emph{NeuroImage}, vol.~46, no.~4, pp.
  1066--1081, July 2009.

\bibitem{haufe:09}
S.~Haufe, K.~M{\"{u}}ller, G.~Nolte, and N.~Kr{\"{a}}mer, ``Sparse causal
  discovery in multivariate time series,'' \emph{Journal of Machine Learning
  Research: Workshops \& Conference Proceedings (JMLR W\&CP)}, vol. 6 (NIPS
  2008), pp. 97--106, 2010.

\bibitem{lozano:bio09}
A.~Lozano, N.~Abe, Y.~Liu, and S.~Rosset, ``Grouped graphical granger modeling
  for gene expression regulatory networks discovery,'' \emph{Bioinformatics},
  vol.~25, pp. i110 -- i118, 2009.

\bibitem{vandeGeer:09}
S.~van~de Geer and P.~B{\"{u}}hlmann, ``On the conditions used to prove oracle
  results for the lasso,'' \emph{Electronic Journal of Statistics}, vol.~3, pp.
  1360--1392, 2009.

\bibitem{wright:08}
S.~Wright, R.~Nowak, and M.~Figueiredo, ``Sparse reconstruction by separable
  approximation,'' in \emph{IEEE International Conference on Acoustics, Speech,
  and Signal Processing}, Las Vegas, NV, 2008.

\bibitem{lozano:nips09}
A.~Lozano, G.~Swirszcz, and N.~Abe, ``Grouped orthogonal matching pursuit for
  variable selection and prediction,'' in \emph{Advances in Neural Information
  Processing Systems (NIPS)}, no.~22, 2009.

\bibitem{meins:06}
N.~Meinshausen and P.~B{\"{u}}hlmann, ``High-dimensional graphs and variable
  selection with the lasso,'' \emph{The Annals of Statistics}, vol.~34, no.~3,
  pp. 1436--1462, 2006.

\bibitem{ravikumar:08}
P.~Ravikumar, G.~Raskutti, M.~Wainwright, and B.~Yu, ``Model selection in
  gaussian graphical models: High-dimensional consistency of
  $\ell_1$-regularizedmle,'' in \emph{Advances in Neural Information Processing
  Systems (NIPS)}, no.~21, 2008.

\bibitem{friedman:08}
J.~Friedman, T.~Hastie, and R.~Tibshirani, ``Sparse inverse covariance
  estimation with the graphical lasso,'' \emph{Biostatistics}, vol.~9, no.~3,
  pp. 432--441, 2008.

\bibitem{lozano:kdd09}
A.~Lozano, N.~Abe, Y.~Liu, and S.~Rosset, ``Grouped graphical granger modeling
  methods for temporal causal modeling,'' 2009, pp. 577 -- 585.

\bibitem{zou:06}
H.~Zou, ``The adaptive lasso and its oracle properties,'' \emph{Journal of the
  American Statistical Association}, vol. 101, pp. 1418--1429, 2006.

\bibitem{tropp_ssac:06}
J.~Tropp, A.~Gilbert, and M.~Strauss, ``Algorithms for simultaneous sparse
  approximation. part ii: Convex relaxation,'' \emph{Signal Processing, special
  issue on Sparse approximations in signal and image processing}, vol.~86, pp.
  589--602, April 2006.

\bibitem{meier:08}
L.~Meier, S.~van~de Geer, and P.~B{\"{u}}hlmann, ``The group lasso for logistic
  regression,'' \emph{Journal of the Royal Statistical Society Series B},
  vol.~70, no.~1, pp. 53--71, 2008.

\bibitem{cotter:05}
S.~Cotter, B.~Rao, K.~Engan, and K.~Kreutz-Delgado, ``Sparse solutions to
  linear inverse problems with multiple measurement vectors,'' \emph{IEEE
  Transactions on Signal Processing}, vol.~53, no.~7, pp. 2477--2488, July
  2005.

\bibitem{chen_huo:04}
J.~Chen and X.~Huo, ``Theoretical results on sparse representations of
  multiple-measurement vectors,'' \emph{IEEE Transactions on Signal
  Processing}, vol.~54, no.~12, pp. 4634--4643, December 2006.

\bibitem{tropp_ssag:06}
J.~Tropp, A.~Gilbert, and M.~Strauss, ``Algorithms for simultaneous sparse
  approximation. part i: Greedy pursuit,'' \emph{Signal Processing, special
  issue on Sparse approximations in signal and image processing}, vol.~86, pp.
  572--588, April 2006.

\bibitem{obozinski:08}
G.~Obozinski, M.~Wainwright, and M.~Jordan, ``Union support recovery in
  high-dimensional multivariate regression,'' UC Berkeley, Tech. Rep. 761,
  August 2008.

\bibitem{wang_leng:08}
H.~Wang and C.~Leng, ``A note on adaptive group lasso,'' \emph{Computational
  Statistics and Data Analysis}, vol.~52, pp. 5277�--5286, 2008.

\bibitem{liu_zhang:09}
H.~Liu and J.~Zhang, ``Estimation consistency of the group lasso and its
  applications,'' \emph{Journal of Machine Learning Research: Workshops \&
  Conference Proceedings (JMLR W\&CP)}, vol.~5, pp. 376--383, 2009.

\bibitem{valdes:05}
P.~Valdes-Sosa, J.~Sanchez-Bornot, J.~Lage-Castellanos, M.~Vega-Hernandez,
  J.~Bosch-Bayard, L.~Melie-Garcia, and E.~Canales-Rodriguez, ``Estimating
  brain functional connectivity with sparse multivariate autoregression,''
  \emph{Philosophical Transactions of the Royal Society B}, vol. 360, pp.
  969--981, 2005.

\bibitem{songsiri:10}
J.~Songsiri and L.~Vandenberghe, ``Topology selection in graphical models of
  autoregressive processes,'' \emph{J. Machine Learning Research}, no.~11, pp.
  2671--2705, 2010.

\bibitem{bento:10}
J.~Bento, M.~Ibrahimi, and A.~Montanari, ``Learning networks of stochastic
  differential equations,'' \emph{{\tt arXiv:1011.0415v1 [math.ST]}}, 2010.

\bibitem{meins:10}
N.~Meinshausen and P.~B{\"{u}}hlmann, ``Stability selection,'' \emph{J. Royal
  Statistical Society B}, vol.~72, no.~4, pp. 417--473, 2010.

\bibitem{pereda:05}
E.~Pereda, R.~Q. Quiroga, and J.~Bhattacharya, ``Nonlinear multivariate
  analysis of neurophysiological signals,'' \emph{Progress in Neurobiology},
  vol.~77, no.~1, pp. 1--37, 2005.

\bibitem{carroll:09}
M.~Carroll, G.~Cecchi, R.~Rish, R.~Garg, and A.~Rao, ``Prediction and
  interpretation of distributed neural activity with sparse models,''
  \emph{NeuroImage}, vol.~44, no.~1, pp. 112--122, January 2009.

\bibitem{haufe:09b}
S.~Haufe, R.~Tomioka, G.~Nolte, K.~M{\"{u}}ller, and M.~Kawanabe, ``Modeling
  sparse connectivity between underlying brain sources for eeg/meg,''
  \emph{IEEE Transactions on Biomedical Engineering}, vol.~57, no.~8, pp.
  1954--1963, 2010.

\bibitem{young:93}
M.~Young, ``The organization of neural systems in the primate cerebral
  cortex,'' \emph{Proc. Biol. Sci.}, no. 252, pp. 13--18, 1993.

\bibitem{sporns:02}
O.~Sporns, \emph{Graph theory methods for the analysis of neural connectivity
  patterns}, R.~K{\"{u}}tter, Ed.\hskip 1em plus 0.5em minus 0.4em\relax
  Boston: Kl{\"{u}}wer, 2002.

\bibitem{hothorn:08}
T.~Hothorn, F.~Bretz, and P.~Westfall, ``Simultaneous inference in general
  parametric models,'' \emph{Biometrical Journal}, vol.~50, pp. 346 -- 363,
  2008.

\bibitem{watts:98}
D.~Watts and S.~Strogatz, ``Collective dynamics of `small-world' networks,''
  \emph{Letters to Nature}, vol. 393, pp. 440--442, June 1998.

\bibitem{sporns:07}
O.~Sporns, C.~Honey, and R.~K{\"{o}}tter, ``Identification and classification
  of hubs in brain networks,'' \emph{PLoS ONE}, vol.~2, p. e1049, October 2007.

\bibitem{sporns:00}
O.~Sporns, G.~Tononi, and G.~Edelman, ``Theoretical neuroanatomy: Relating
  anatomical and functional connectivity in graphs and cortical connection
  matrices,'' \emph{Cerebral Cortex}, vol.~10, pp. 127 -- 141, 2000.

\bibitem{laurent:00}
B.~Laurent and P.~Massart, ``Adaptive estimation of a quadratic function by
  model selection,'' \emph{Annals of Statistics}, vol.~28, no.~5, pp.
  1302�--1338, 2000.

\end{thebibliography}

\end{document}